\documentclass[sigconf, nonacm]{lib/acmart}

\AtBeginDocument{%
  \setlength{\abovedisplayskip}{0pt}      
  \setlength{\belowdisplayskip}{0pt}      
}

\usepackage{tikz}
\usepackage{amsmath}

\usepackage[linesnumbered, ruled]{algorithm2e}

\usepackage{colortbl}
\usepackage{eucal}
\usepackage{bm}
\usepackage{booktabs}
\usepackage{multirow}
\usepackage{enumitem}
\usepackage{url}
\usepackage{xspace}
\usepackage{amsfonts}
\usepackage[english]{babel}
\usepackage{blindtext}

\usepackage{pifont} 
\usepackage{stackengine} 

\usepackage{setspace}

\usepackage{siunitx}

\usepackage{threeparttable} 
\usepackage{tabularx}

\usepackage{caption}
\usepackage{subcaption}
\usepackage{hyperref}

\usepackage{booktabs}

\newcommand{\cmark}{\ding{51}}  
\newcommand{\xmark}{\ding{55}}  

\newcolumntype{C}[1]{>{\centering\arraybackslash}p{#1}}
\newcolumntype{L}[1]{>{\raggedright\arraybackslash}p{#1}}

\def\eg{\textit{e.g.},\xspace}
\def\etal{\textit{et al.}\xspace}

\newcommand{\vsnoindent}{} 

\newcommand{\sysname}{\texttt{Carbon}\emph{X}\xspace}

\newcommand{\equal}{\textsuperscript{\dag}}   

\begin{document}

\title{\sysname: An Open-Source Tool for Computational Decarbonization Using Time Series Foundation Models}

\author{%
  Diptyaroop Maji\textsuperscript{1}\equal, 
  Kang Yang\textsuperscript{2}\equal, 
  Prashant Shenoy\textsuperscript{1}, 
  Ramesh K. Sitaraman\textsuperscript{1}, 
  Mani Srivastava\textsuperscript{2}
}

\authornote{Mani Srivastava holds concurrent appointments as a Professor of ECE and CS (joint) at the University of California, Los Angeles, and as an Amazon Scholar at Amazon. This paper describes work performed at UCLA and is not associated with Amazon.}

\affiliation{%
  \institution{\textsuperscript{1}University of Massachusetts Amherst \quad \textsuperscript{2}University of California Los Angeles}
  \country{}
}

\renewcommand{\shortauthors}{Maji, Yang et al.}

\begin{abstract}


Computational decarbonization aims to reduce carbon emissions in computing and societal systems such as data centers, transportation, and built environments.  
This requires accurate, fine-grained carbon intensity forecasts, yet existing tools have several key limitations: (i) they require grid-specific electricity mix data, restricting use where such information is unavailable; (ii) they depend on separate grid-specific models that make it challenging to provide global coverage; and (iii) they provide forecasts without uncertainty estimates, limiting reliability for downstream carbon-aware applications.

In this paper, we present \sysname, an open-source tool that leverages Time Series Foundation Models (TSFMs) for a range of decarbonization tasks. \sysname utilizes the versatility of TSFMs to provide strong performance across multiple tasks, such as carbon intensity forecasting and imputation, and across diverse grids. 
Using only historical carbon intensity data and a single general model, our tool achieves a zero-shot forecasting Mean Absolute Percentage Error (MAPE) of 15.82\% across 214 grids worldwide.
Across 13 benchmark grids, \sysname performance is comparable with the current state-of-the-art, with an average MAPE of 9.59\% and tail forecasting MAPE of 16.54\%, while also providing prediction intervals with 95\% coverage. \sysname can provide forecasts for up to 21 days with minimal accuracy degradation.
Further, when fully fine-tuned, \sysname outperforms the statistical baselines by 1.2---3.9$\times$ on the imputation task. Overall, these results demonstrate that \sysname can be used easily on any grid with limited data and still deliver strong performance, making it a practical tool for global-scale decarbonization.

\end{abstract}


\maketitle

\begingroup
\renewcommand\thefootnote{\dag}%
\footnotetext{Both authors contributed equally to this research.}%
\endgroup

\sloppy

\section{Introduction}\label{sec_introduction}

Societal infrastructure, including electricity grids, buildings, transportation, and computing systems, contributes over~80\% of global greenhouse gas~(GHG) emissions, underscoring the need for scalable decarbonization~\cite{gas_epa}.
To reduce such GHG emissions, a new field of computational decarbonization is gaining attention in recent years~\cite{irwin2025vision}, which aims to reduce both operational and embodied carbon emissions associated with complex computing and societal infrastructure systems by leveraging sensing, optimization, and Artificial Intelligence~(AI)-driven control.

A key enabler for computational decarbonization is accurate, fine-grained estimates and forecasts of the electricity grid's \emph{carbon intensity}, which quantifies the emissions associated with per unit electricity usage over a period (in~\si{gCO_2eq/kWh}). 
Carbon intensity depends on several factors such as the electricity (energy) mix, weather, and demand patterns~\cite{maji2022carboncast}, and therefore, varies significantly both spatially and temporally. 
For example, Figure~\ref{fig_year2020} shows the carbon intensities in two locations: California exhibits a ``duck curve'' with midday solar dips and evening gas spikes, while Germany shows a higher and more volatile pattern driven by fossil fuels and intermittent wind or solar generation. 
Systems can take advantage of such variability to reduce their carbon emissions without compromising performance~\cite{irwin2025vision, thiede2023carbon, carbon-explorer, waitawhile, majibringing}. 
For instance, flexible demands such as electric vehicle charging or large-scale Machine Learning~(ML) model training can be scheduled during low-carbon periods to reduce the associated emissions. 
Thus, fine-grained carbon intensity estimates and accurate forecasts, when accessible at a global scale, can help in a significant reduction of carbon emissions and subsequently advance the decarbonization goals.

{\bf Limitations of Prior Work.}
In recent years, several ML-based specialized tools have emerged that can provide accurate short-term grid carbon intensity forecasts~\cite{maji2022dacf, maji2022carboncast, yan2025ensembleci, zhang2023gnn}. 
Although these tools show considerable promise, these tools face three key limitations when availability at a global scale is considered:

\begin{figure}[t]
\centering
{\includegraphics[width=0.85\linewidth]{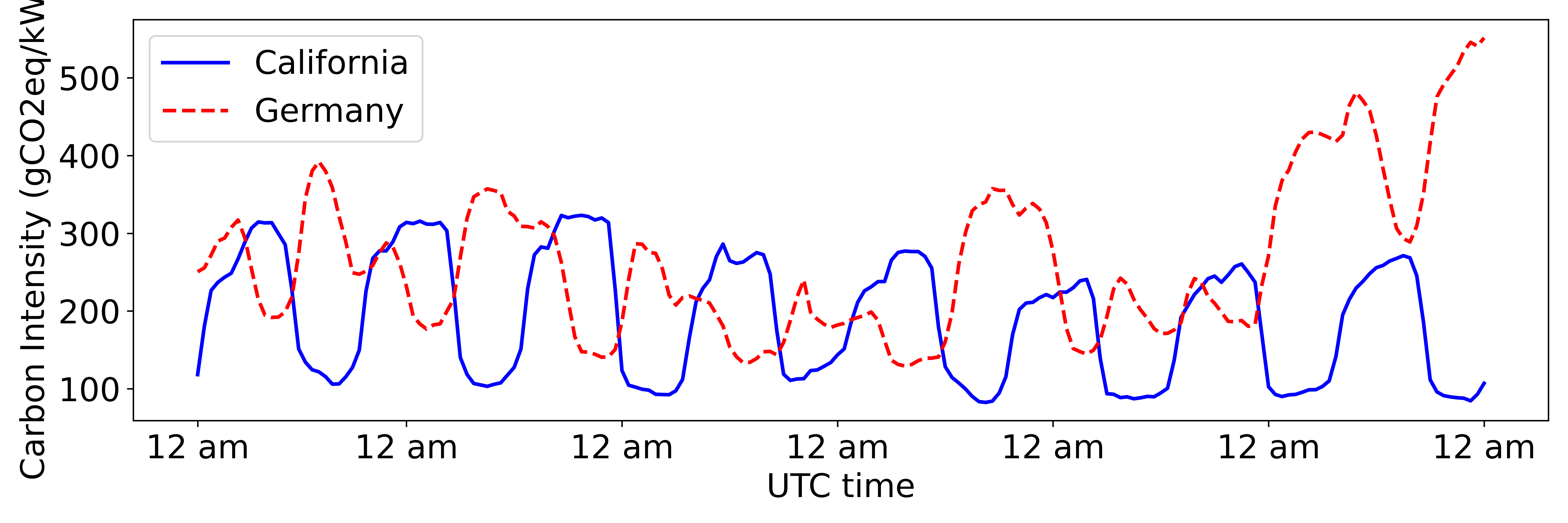}}
\caption{\emph{Hourly carbon intensity~(\si{gCO_2eq/kWh}) in the California and Germany grids over one week.}}
\Description[]{}
\label{fig_year2020}
\end{figure}

\noindent
\emph{(i) Dependence on data that are not generally available.} 
Existing tools rely on information about a grid’s source mix to generate accurate carbon intensity forecasts.  
While some grid operators~\cite{caiso, eia-grid-monitor} and services~\cite{electricitymaps, watttime} have begun releasing historical carbon intensity data, reporting source mix remains non-mandatory~\cite{ghg_scope2} and is thus unavailable for many grids, limiting the applicability of existing models. For example, there are entire continents, such as Africa, where electricity mix data is either not available or only available at an annual scale~\cite{iea-africa}; hence, prior tools cannot be used.

\noindent
\emph{(ii) Dependence on grid-specific models.}  
Prior tools rely on grid-specific models trained separately for each individual grid, and on some occasions, even individual electricity sources~\cite {maji2022carboncast}. Although this design effectively captures grid-specific carbon dynamics, it is highly challenging to scale globally across grids. For example, at the transmission system operator level, there are more than 350 grids worldwide~\cite{emaps-datasets}. With each grid having multiple electricity sources, prior works need to train and maintain thousands of models. At the distribution level, there are more than 2500 operators alone in Europe~\cite{eu-dso}, making the per-grid design even more challenging.

\noindent
\emph{(iii) No quantification of forecast uncertainty.}  
Similar to other forecasting systems~(\eg weather forecasts), carbon intensity prediction involves inherent uncertainty.  
Most prior works generate point forecasts without capturing possible variations.  
Such forecasts can deviate significantly from the ground truth, especially over multi-day horizons~\cite{li2024uncertainty}, reducing the reliability of downstream applications that perform optimizations by relying on these forecasts.

{\bf Our Goal.}
These gaps motivate the need for a tool that ~(i)~operates without relying on auxiliary information such as energy mix, 
(ii) generalizes across global electricity grids using a unified architecture, 
(iii) produces uncertainty-aware forecasts suitable for real-world deployment, and is also easy to use and maintain at a global scale.  
In addition, tasks such as carbon intensity imputation remain underexplored despite their importance for carbon-aware optimizations. For example, imputation can recover missing data or enhance temporal resolution in regions with sparse carbon intensity reporting~\cite{electricitymaps}, increasing the effectiveness of downstream carbon-aware applications.
Therefore, there is also a need for a tool that can perform other tasks, such as carbon intensity imputation, in addition to forecasting.

Lately, large-scale ML models, referred to as Time Series Foundation Models~(TSFMs)~\cite{goswami_moment_2024, ansari2024chronos, das_decoder-only_2024, liu2025sundial, shi2024time}, are showing promise in forecasting and imputation tasks on time series data from other domains such as electricity and weather~\cite{goswami_moment_2024, ansari2024chronos}. 
Pre-trained on diverse time-series data across multiple domains, TSFMs capture rich temporal patterns through a unified architecture and often have strong zero- and few-shot performance on unseen domains using only an autoregressive approach. Thus, a tool that leverages the capabilities of TSFMs while operating on carbon intensity data can potentially overcome the limitations of the prior tools and meaningfully advance computational decarbonization.



\textbf{Our Research Contributions.} In this paper, we present \sysname, a tool built using TSFMs for tasks such as carbon intensity forecasting and imputation.  
Using only carbon intensity series and a single general model, \sysname achieves an average Mean Absolute Percentage Error~(MAPE) of 15.82\% across 214 grids worldwide for the forecasting task.
Although \sysname has slightly higher forecasting errors than existing specialized tools in some grids, it delivers similar tail performance (see Section~\ref{sec:sota-comparison}) using only historical carbon intensity as input. 
Further, it offers significantly more flexibility and versatility (see Table~\ref{tab:comparing-sota-with-our-work}), such as one model for any grid and tasks beyond forecasting. These attributes make \sysname more practical than any similar tools existing today.
In summary, this paper makes the following contributions:


\begin{enumerate}[leftmargin=*, topsep=2pt]

\item \emph{Open-Source Tool.}  We present \sysname, an open source tool for global carbon intensity forecasting and imputation. \sysname has a modular design that leverages the versatility of TSFMs to provide comparable accuracy with the current state-of-the-art using only historical carbon intensity data as input. We release \sysname with the saved models, curated datasets, and an API suite to facilitate decarbonization research. These resources are available at \textbf{\url{https://github.com/codecexp/CarbonX}}.

\item \emph{Global Coverage with Strong Forecasting Accuracy.} \sysname provides multi-day carbon intensity forecasts across 214 grids globally. Due to its zero-shot capability, which requires no additional training, support for any new grids can also be added easily. On the 13 benchmark grids used in prior works, \sysname attains 9.59\% mean MAPE and 16.54\% tail MAPE, compared to 7.75\% and 15.61\% for the current state-of-the-art. Across all 214 grids, the mean and tail MAPEs are 15.82\% and 29.61\%, respectively. \sysname is the first research work to provide and evaluate carbon intensity forecasting on a global scale.

\item \emph{Extended Forecasting Horizon.}  
\sysname can forecast carbon intensity for an extended horizon with minimal performance loss. Our evaluation across 13 grids shows that \sysname can forecast up to 21 days ahead with only a 5.40\% average MAPE degradation.  
To our knowledge, it is the first work to support carbon intensity forecasting beyond a 96-hour horizon.

\item \emph{Uncertainty Quantification.}  
\sysname provides calibrated Prediction Intervals (PIs) to make the outputs reliable for downstream tasks. For forecasting, it achieves 95\% average coverage with a 54.2\% normalized interval width across 13 grids, demonstrating effective uncertainty quantification.

\item \emph{Imputation Support.}  
\sysname also supports carbon intensity imputation, offering versatility beyond forecasting. Our fully fine-tuned implementation achieves an average normalized RMSE of 0.24 across 50 grids even with 75\% missing values, outperforming the statistical baselines by 1.2---3.9$\times$.

\end{enumerate}

\section{Background}
\label{sec_background}


\subsection{Electricity Grids and Carbon Intensity}

The electricity grid combines generation from renewable and non-renewable sources.  
Electricity is transmitted through high-voltage lines and distributed to end users.  
Each regional grid is managed by an operator who balances supply and demand.  
As demand varies, operators monitor the grid and weather conditions in real time, and dispatch generators accordingly to meet current demand.


\begin{table}[t!]
\centering
\resizebox{0.95\columnwidth}{!}{
\begin{tabular}{lcc}
\toprule
\textbf{TSFM} & \textbf{Variant} & \textbf{Parameters} \\
\midrule

MOMENT~\cite{goswami_moment_2024} & Large (\textit{AutonLab/MOMENT-1-large}) & 385M \\

Chronos~\cite{ansari2024chronos} & Large (\textit{amazon/chronos-t5-large}) & 710M \\

TimesFM~\cite{das_decoder-only_2024} & \textit{TimesFM-2.0-500M} & 500M \\

Sundial~\cite{liu2025sundial} & \textit{thuml/sundial-base-128m} & 128M \\

Time-MoE~\cite{shi2024time} & \textit{Maple728/TimeMoE-50M} & 50M \\


\bottomrule
\end{tabular}
}
\caption{\emph{Supported TSFMs with variants and parameter sizes.}}
\label{tab:tsfm-evaluated}
\end{table}

\begin{table*}[t!]
\centering
\begin{threeparttable}
\resizebox{0.9\textwidth}{!}{
\begin{tabular}{l|cccccc}
\toprule

\textbf{} & \textbf{Energy mix} & \textbf{Training data volume} & \textbf{Transferrable} & \textbf{Prediction} & \textbf{Tasks beyond} & \textbf{Forecasting}\\

\textbf{} & \textbf{not required} & \textbf{\& training frequency} & \textbf{across grids} & \textbf{intervals} & \textbf{forecasting} & \textbf{accuracy} \\

\midrule

EWMA~\cite{shenoy2024home} & \cmark & High  & \xmark & \xmark & \xmark & Moderate \\

CarbonCast~\cite{maji2022carboncast} & \xmark & High & \xmark & \xmark & \xmark & High\\

CarbonCast w/ PI~\cite{li2024uncertainty} & \xmark & High & \xmark & \cmark & \xmark & High\\

EnsembleCI~\cite{yan2025ensembleci} & \xmark & High & \xmark & \xmark & \xmark & High\\

\rowcolor{gray!20} \textbf{\sysname} & \textbf{\cmark} & \textbf{Low} & \textbf{\cmark} & \textbf{\cmark} & \textbf{\cmark} & \textbf{Moderate-High} \\

\bottomrule
\end{tabular}
}
\end{threeparttable}
\caption{\emph{Comparison between \sysname and existing state-of-the-art models.}}
\vspace{-0.1in}
\label{tab:comparing-sota-with-our-work}
\end{table*}

\noindent
\textbf{Carbon Intensity~(CI).}
The average carbon intensity of a grid~(in~\si{gCO_2eq/kWh}) is the generation-weighted average of emissions from each electricity source at a certain time:
\begin{equation}
    \text{Carbon Intensity~(CI)} = \frac{\sum (E_i \cdot CEF_i)}{\sum E_i} ,
\end{equation}
where~$E_i$ is the electricity generated by source~$i$~(\eg coal, solar, gas) in~\si{kWh}, and~$CEF_i$ is its corresponding carbon emission factor~(CEF, in~\si{gCO_2eq/kWh}), a standardized value for each source~\cite{schlomer2014annex, greenhouse_gas}.

\subsection{Time Series Foundation Models (TSFMs)}

A TSFM is a large-scale ML model pre-trained on diverse time-series datasets~\cite{goswami_moment_2024, ansari2024chronos}.  
By learning universal temporal representations, it enables zero-shot or few-shot adaptation across domains.
TSFMs have been applied to multiple domains ranging from electricity load forecasting to healthcare~\cite{asgharnezhad2024time, liang2024enabling, goel2025foundation, nguyen2023climax, kalahasti2025foundation}.
These applications demonstrate TSFMs' ability to generalize across domains and encourage application of new or existing TSFMs on carbon intensity data.

\subsubsection{\textbf{Supported TSFMs.}}
Table~\ref{tab:tsfm-evaluated} lists the TSFM variants currently supported by \sysname, and their parameter counts.  
These models differ in architecture~(\eg encoder-decoder) and training strategy~(\eg supervised, self-supervised).
The modular design of \sysname enables seamless integration of current and future TSFMs.

\noindent
\textbf{MOMENT}~\cite{goswami_moment_2024}: 
A transformer-based model, pre-trained using a masked time series. 
It segments input sequences into fixed-length patches and encodes them using a modified T5-style~\cite{raffel2020exploring} transformer.
A reconstruction head then enables patch-level prediction. 

\noindent
\textbf{Chronos}~\cite{ansari2024chronos}:
A transformer-based forecasting model that repurposes language models for time series.
It tokenizes time series via scaling and quantization into a fixed-size vocabulary, enabling training with off-the-shelf encoder-decoder~\cite{raffel2020exploring}.
It uses cross-entropy loss to model the categorical distribution of tokenized values.


\noindent
\textbf{TimesFM}~\cite{das_decoder-only_2024}: A decoder-only model that processes time series as sequences of non-overlapping patches, which are embedded using residual multilayer perceptron blocks and passed through stacked transformer layers with causal self-attention. 

\noindent
\textbf{Sundial}~\cite{liu2025sundial}: A Transformer-based model that pre-trains directly on continuous time series without tokenization, leveraging a flow-matching TimeFlow loss to predict next-patch distributions, mitigate mode collapse, and generate arbitrary-length forecasts.

\noindent
\textbf{Time-MoE}~\cite{shi2024time}: A decoder-only Transformer with a sparse Mixture-of-Experts (MoE) design that activates only a few experts per step, reducing computation while preserving model capacity. Similar to others, it supports variable forecasting horizons.

\subsubsection{\textbf{TSFM Fine-Tuning.}}
TSFMs show improved performance under a few-shot setting, when fine-tuned with data from the same domain on which they are applied~\cite{goswami_moment_2024}. 
Fine-tuning strategies can be categorized as \emph{model-oriented} or \emph{data-oriented}.

\vsnoindent
In \emph{model-oriented fine-tuning}, either a subset or the entire model is updated using domain-specific data.  
Fine-tuning only part of the model, \eg the task head layers, is referred to as \emph{lightweight} fine-tuning. 
Updating all model parameters constitutes \emph{full} fine-tuning.

\vsnoindent
\emph{Data-oriented fine-tuning} for grid data falls into two types.
In \emph{domain-specific} fine-tuning, the model is trained on data from multiple grids and evaluated on an unseen grid, testing inter-grid generalization.  
In \emph{grid-specific} fine-tuning, the model is fine-tuned and evaluated on data from the same grid, testing intra-grid accuracy.

\section{Related Work}

Table~\ref{tab:comparing-sota-with-our-work} summarizes the key differences and advantages of our tool relative to previous approaches.

\subsection{\textbf{Carbon Intensity Forecasting}}
Short-term carbon intensity forecasting~(up to 96 hours) is increasingly well-studied due to its importance for decarbonization.  
Traditional time-series forecasting methods, such as EWMA~\cite{shenoy2024home} and ARIMA~\cite{seabold2010statsmodels, bokde2021short, leerbeck2020short}, have been applied in several prior works because of their simplicity and effectiveness in capturing temporal patterns.  
However, such statistical models struggle to handle high variability and are less effective for multi-day forecasting horizons.

\noindent
\textbf{Domain-Specific ML Models.}  
To address the limitations of traditional statistical methods, ML models have emerged as state-of-the-art solutions~\cite{ maji2022carboncast, zhang2023gnn, yan2025ensembleci}.  
CarbonCast~\cite{maji2022carboncast} employs a two-tier ML architecture for 96-hour carbon intensity forecasting.  
The first tier predicts electricity generation for each energy source. The second tier takes these predictions along with historical carbon intensity and weather forecasts to estimate 96-hour carbon intensity forecasts.
More recently, EnsembleCI~\cite{yan2025ensembleci} applies ensemble learning across three base models to produce 96-hour forecasts.  
Each base model predicts the carbon intensity of a given grid, and a stacking-based strategy is used to generate the final forecast. 
Despite their effectiveness, these models rely on extensive grid-specific electricity mix data, frequent retraining, and grid-specific model development to maintain accuracy.
In contrast, \sysname requires only historical carbon intensity and achieves similar accuracy in many grids. Additionally, \sysname also provides prediction intervals. 

\noindent
\textbf{Proprietary Forecasting Services.}
Electricity Maps~\cite{electricitymaps} and WattTime~\cite{watttime} offer global carbon intensity forecasts.
However, these forecasts are proprietary, and their underlying models and grid-wise forecasting accuracy are not publicly available.  
In contrast, our tool is fully open-sourced to support transparency and reproducibility for researchers and practitioners.

\noindent
\textbf{Forecast Uncertainty.}
Most prior works provide only point-value forecasts and do not quantify uncertainty.
Recently, Li~\etal~\cite{li2024uncertainty} have extended CarbonCast~\cite{maji2022carboncast}, where they apply a sequential conformal prediction method~\cite{xu2023sequential} to generate prediction intervals. However, underlying limitations remain, such as its reliance on electricity mix data and the per-grid model.
In contrast, \sysname requires only historical carbon intensity data to generate intervals.

\begin{figure}[t]
\centering
{\includegraphics[width=0.85\linewidth]{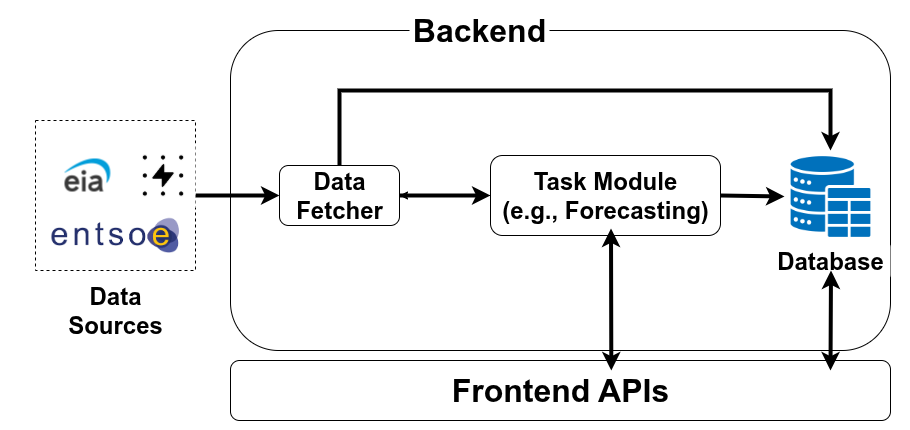}}
\caption{\emph{\sysname end-to-end pipeline. The frontend communicates with the task module and the database during execution.}}
\Description[]{}
\label{fig:codecfm-pipeline}
\end{figure}

\subsection{\textbf{Carbon Intensity Imputation}}
Carbon intensity imputation enables recovery of missing data and generation of higher-resolution estimates than those currently available. For example, given hourly carbon intensity data, imputation can be applied to generate sub-hourly values. 
At present, only Electricity Maps performs imputation using a combination of statistical and machine learning techniques~\cite{emaps-data-estimation}. However, their models are proprietary. 
While statistical methods are commonly used for data imputation, to the best of our knowledge, \sysname is among the first to explore carbon intensity imputation as a standalone task.

\section{\texttt{C\MakeLowercase{arbon}}\emph{X} design}
\label{sec_design}

\sysname is a generalizable, versatile tool that performs carbon domain tasks across electricity grids with high accuracy, using univariate carbon intensity series.  
It currently supports multi-day forecasting and imputation, and can be extended to other decarbonization tasks, such as anomaly detection, with minimal changes.

\subsection{End-to-end Pipeline and Usage}  

We begin by briefly describing the end-to-end pipeline and illustrating how to use \sysname.
Figure~\ref{fig:codecfm-pipeline} illustrates the \sysname{} pipeline, including the frontend and backend, where the frontend interacts with the backend’s task module and database during execution.

\subsubsection{\textbf{\sysname Backend.}}
\sysname backend comprises a data fetcher module, a task module, and a database to store both fetched and generated data.

\noindent\textbf{Data Fetcher.} The data fetcher module periodically retrieves carbon intensity data from various public data sources. Currently, \sysname fetches hourly carbon intensity data from Electricity Maps~\cite{emaps-datasets} for 214 grids. We plan to include other sources, such as EIA~\cite{eia-grid-monitor} and ENTSOE~\cite{entsoe-data}, in the future. While the module currently runs once daily, the periodicity can be adjusted as per requirements. Once fetched, the data is forwarded to the task module as well as updated to the database.

\noindent\textbf{Task Module.} 
The task module receives data from the data fetcher and performs the desired task. Regardless of the task, it first runs imputation (using MOMENT) so that any missing data is filled.

\begin{figure}[t]
\centering
{\includegraphics[width=\columnwidth]{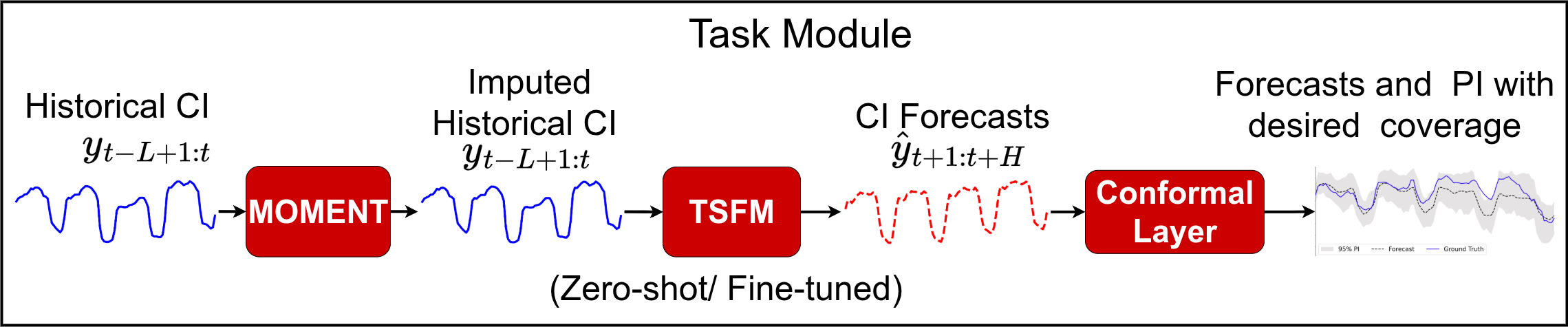}}
\caption{\emph{\sysname task module comprises TSFMs followed by a conformal layer. This figure shows the forecasting task. The TSFM uses $L$ timesteps of historical data to forecast the next $H$ timesteps.
A conformal layer then produces calibrated prediction intervals with desired coverage.}}
\Description[]{}
\label{fig:cifm-design}
    \vspace{0.1in}
\end{figure}

\vsnoindent
Figure~\ref{fig:cifm-design} illustrates the \sysname task module for carbon intensity forecasting. The module can use any new or existing TSFM to forecast~$\hat{y}_{t+1:t+H}$ from historical input~$y_{t-L+1:t}$, where~$L$ is the input length and $H$ is the forecast horizon.  
The output is then passed through a conformal prediction layer, which produces a prediction interval with coverage guarantees. Once the forecasts and prediction intervals are generated, we update the database. We describe the task module in more detail in Section~\ref{sec:task-module}. 

\noindent\textbf{Database.} The database is currently a list of files containing the actual carbon intensity data, carbon intensity forecasts, and prediction intervals for all the supported grids. The database is updated daily after the data fetcher and task modules run.

\subsubsection{\textbf{\sysname Frontend.}} 
The frontend provides a suite of APIs that users or downstream services can use to easily access any historical carbon intensity data, as well as forecasts from any prior or current date for any supported grid. For example, suppose a data center in Texas requires carbon intensity forecasts for the next day in order to optimally schedule a batch job. Calling \texttt{get\_ci\_forecasts(ERCOT, <date>, 24, True)} will retrieve day-ahead carbon intensity forecasts from \sysname for the Texas grid (ERCOT) along with prediction intervals. 

\vsnoindent
Additionally, \sysname users can set any preferred TSFM and the mode of operation (zero-shot or fine-tuned), query which grids are supported, and even generate finer resolution carbon intensity data as per requirements. Table~\ref{tab:carbonx-api} in Appendix~\ref{appendix:api-suite} provides a description of the APIs currently supported by \sysname, the required parameters, and the values returned. A detailed documentation can be found in \textbf{\url{https://github.com/codecexp/CarbonX}}.

\subsection{\sysname Task Module}
\label{sec:task-module}
The \sysname task module takes only a univariate carbon intensity time series as input and performs forecasting or imputation. The task module consists of three sub-modules:

\vsnoindent(i) Regardless of any task, the input to the task module is passed through a TSFM that performs carbon intensity imputation. This step ensures that any missing data is accurately estimated before further processing. The TSFM can operate in zero-shot or fine-tuned mode per requirements. At present, \sysname uses MOMENT for imputation. If the desired task is imputation, the next sub-module is skipped, and the data is sent to the conformal layer.

\vsnoindent(ii) For other tasks, the data is input to a second TSFM. This sub-module is model-agnostic and can leverage any TSFM best suited for the task, providing flexibility to evaluate both newly developed and existing models, including those from other domains. For instance, we plug in five representative TSFMs to assess their forecasting performance. The TSFMs can operate either in zero-shot or fine-tuned modes. The output is passed to the conformal layer.

\vsnoindent(iii) The final sub-module is a conformal layer that provides calibrated intervals with desired coverage given a point-value time series, where coverage is defined as the percentage of times the actual values fall within the interval.
Tasks such as multi-day forecasting involve significant uncertainty, as forecasts get progressively worse with an increasing horizon. Many existing TSFMs have the capability to provide probabilistic forecasts from which prediction intervals can be obtained. 
However, providing intervals naively without any coverage guarantees may not be helpful. To address this, \sysname applies a conformal prediction technique on the point-value data. The conformal layer calibrates the intervals to \emph{cover} the ground truth 95\% of the time.

Our modular design ensures that any TSFM or any conformal prediction technique best suited for a particular task can be applied with minimal changes. This enhances the accuracy and reliability of \sysname, making it a reliable tool for downstream decarbonization applications that require robust carbon intensity estimates.

\vsnoindent
Next, we describe the zero-shot and fine-tuned modes of operation in detail, followed by our uncertainty quantification approach.

\subsubsection{\textbf{Zero-Shot Mode.}}

There are three challenges when applications attempt to estimate or forecast carbon intensity to assess their carbon emissions.  
First, many power grids lack information beyond historical carbon intensity.  
Second, even when such data is available, it may contain large gaps due to reporting errors or power outages.  
Third, maintaining separate models for each grid becomes increasingly difficult as applications scale across regions.  

\vsnoindent
The \emph{zero-shot} mode addresses these challenges by applying TSFMs pre-trained on data from other domains directly to unseen carbon-related tasks. A single TSFM is deployed in all grids, without any task- or grid-specific fine-tuning. This ``plug-and-play'' mode is useful for short-term tasks where forecasts are only required for a limited amount of time, or in grids where data for fine-tuning is unavailable. The zero-shot mode is also practical in real-world deployment scenarios where global carbon intensity data is required to make optimization decisions. For example, spatial demand-shifting across data centers to reduce carbon emissions may require forecasts from all the grids where the data centers are located. In such scenarios, the zero-shot mode will be beneficial, while maintaining separate models per grid may be infeasible.

\vsnoindent\emph{(i) Zero-shot Forecasting.}
For zero-shot forecasting, \sysname input consists solely of univariate historical carbon intensity data from electricity grids. At present, \sysname does not include any other information that may be available (\eg weather forecasts). However, covariate support can be added with minimal changes. As mentioned, we evaluate five TSFMs in this mode (see Table~\ref{tab:tsfm-evaluated}).

\vsnoindent
During zero-shot forecasting, we also vary the input length~(\eg 24 to 480 hours) and output length~(\eg 1 to 21 days) to study how historical context and forecasting horizon shape model behavior. Short inputs may capture recent trends and support short-term responsiveness, while longer inputs may reveal seasonal patterns but risk introducing noise. On the other hand, TSFMs may be able to generalize over a longer time due to their large model size, contrary to existing domain-specific ML models~\cite{maji2022carboncast, yan2025ensembleci, zhang2023gnn} that typically support only up to 4-day forecasts due to limited model capacity.

\vsnoindent
\emph{(ii) Zero-shot Imputation.}
At present, \sysname uses MOMENT for imputation, since only MOMENT supports imputation among the explored TSFMs. The input for the imputation task is a carbon intensity time series with portions intentionally masked.
This setup mirrors real-world conditions where data gaps arise from reporting delays or power outages~\cite{eia-outage}. Similar to forecasting, a single model is applied uniformly across grids to estimate missing values using the temporal patterns that the model learnt during pre-training.

\subsubsection{\textbf{Fine-Tuned Mode.}} 
Many grids release historical carbon intensity data over extended periods~\cite{emaps-datasets}. Since fine-tuning with data from the target domain typically improves TSFM performance~\cite{goswami_moment_2024}, \sysname supports fine-tuning with domain-specific historical data to provide better carbon intensity forecasts and estimates. This mode can be useful for longer-term deployments or if forecasts are required only for a small number of grids, where it can help capture the temporal distribution in data better than in the zero-shot mode. In this paper, we fine-tune all TSFMs except Sundial, which does not have fine-tuning capabilities yet~\cite{liu2025sundial}.

\vsnoindent
\emph{(i) Fine-Tuned Forecasting.}  
We perform two types of fine-tuning for the forecasting task. First, we employ a \emph{lightweight, grid-specific} fine-tuning strategy to improve the forecasting accuracy. For example, we fine-tune only the linear forecasting head of MOMENT using carbon intensity data ($y$) from a grid while keeping the transformer backbone fixed. At each timestep $t$, the model looks back at the last $L$ time steps ($y_{t-L+1:t}$), and is trained to forecast the next $H$ time steps ($\hat{y}_{t+1:t+H}$). The updated model is then evaluated on unseen data from the same grid.

\vsnoindent
Unlike domain-specific tools such as CarbonCast~\cite{maji2022carboncast} or EnsembleCI~\cite{yan2025ensembleci}, which train separate models from scratch for each grid, we use a shared pre-trained transformer across all grids and update only a few model layers. This enables one-to-few-shot learning and inter-grid generalization, since the majority of the parameters remain unchanged across grids. 

Since the above strategy still needs some grid-specific modifications, we also perform \emph{full,  domain-specific} finetuning to build a scalable, general model. Specifically, we fine-tune all the layers of a TSFM on the combined training data from more than 200 grids. This mode falls in between the zero-shot and grid-specific fine-tuning modes, and is designed to capture diverse carbon intensity patterns without the need for per-grid customization.

Among the TSFMs with fine-tuning capabilities, MOMENT and TimesFM gain primarily from lightweight per-grid adaptation of their forecasting heads and are less suitable for this large-scale strategy. Additionally, Chronos and Time-MoE show little benefit from grid-specific fine-tuning (see Section~\ref{sec:sota-comparison}). Hence, we focus only on Chronos and Time-MoE for the domain-specific strategy.

\vsnoindent\emph{(ii) Fine-Tuned Imputation.} 
We explore both \emph{light-weight} and \emph{full, grid-specific} fine-tuning strategies for the imputation task. The objective is to reconstruct missing carbon intensity values, given a time series. Formally, let y = \(\{y_{1:T}\}\) denote the complete time series over \( T \) time steps, and \( m = \{m_{1:T}\} \) be a binary mask vector, where:
\begin{equation}
m_t =
\begin{cases}
0, & \text{if } y_t \text{ is missing (masked)} \\
1, & \text{if } y_t \text{ is observed}
\end{cases}
\end{equation}

\vsnoindent
The observed data is denoted by \(
y^{\text{obs}} = \{ y_t \mid m_t = 1 \},
\)
and the model is trained to impute the missing values \( \hat{y}_t\ \ \forall t \mid m_t=0\).

\vsnoindent
While the \emph{light-weight} approach emphasizes scalability and generalization, \emph{full fine-tuning} provides high reconstruction accuracy that outperforms the statistical baselines (see Section~\ref{sec:imputation}). We note that \emph{full fine-tuning} results in a separate model for each grid and hence may not always be feasible. However, it achieves strong performance using only historical carbon intensity data and few-shot learning. Hence, we present the results from both strategies to provide a comprehensive comparison, leaving the desired strategy to the downstream application.

\subsubsection{\textbf{Uncertainty Quantification.}}
\label{subsec_uncertainty}

Carbon intensity estimates and forecasts involve considerable uncertainties due to external factors such as weather and changing grid conditions. Given a desired coverage level \( \alpha \in (0, 1) \), \sysname's conformal layer generates prediction intervals that cover the ground truth at least \( 100 \times \alpha\% \) of the time to make the output more robust to uncertainties. For example, \( \alpha = 0.95 \) corresponds to a 95\% coverage. Our approach to ensure the desired coverage is as follows:

\vsnoindent
Since \sysname provides 96-hour forecasts ($\hat{y}$) on a daily basis, historical residuals for hours 25---96 are available after a delay. Suppose $H_{1..96}$ is the forecasting horizon for a day $d$. Then, the most recent historical residuals for $H_{25..48}$ are available from day $d-1$. Similarly, the most recent historical residuals for $H_{73..96}$ are available from day $d-3$. This makes quantifying uncertainty particularly challenging. To quantify effectively and prevent leakage of any future data, we break down the 96-hour forecasting horizon into four forecasting days and compute separate lists of available historical residuals. The conformal layer then looks back at the residuals from the previous available $N$ days to calibrate the current interval. We denote the previous $N$ days as the calibration window, which is a hyperparameter that can be tuned. In our experiments, setting N as 75 provides the best results.

\vsnoindent
We employ a horizon-specific strategy to obtain the prediction intervals. Specifically, for each hour $H$ of the forecasting window, we look at the residual distribution for that particular hour in the calibration window to get the lower ($q^H_{lo} = \alpha/2$) and upper ($q^H_{hi}=1-\alpha/2$) quantiles for that hour. Once we obtain the quantiles, we generate prediction intervals $[L^H, U^H]$ for that hour, where:

\begin{equation}
    L^H = \hat{y^H} + q^H_{lo}, \; U^H = \hat{y^H} + q^H_{hi}
\end{equation}

Thus, the final prediction interval for the whole horizon becomes $[L, U]$, where $L = \hat{y} + q_{lo}, \; U = \hat{y} + q_{hi}$.


\vsnoindent
Since we employ a rolling calibration window for each evaluated day, $q_{lo}$ and $q_{hi}$ reflect the tail behavior of the most recent residuals and provide a data-driven estimate of the uncertainty. Calibrating with the most recent residuals ensures that the intervals adapt to local error patterns and achieve the desired coverage on average.



\section{Experimental Methodology}
\label{sec_methodology}

This section outlines our experimental setup, including datasets, evaluation metrics, TSFM fine-tuning procedures, and the baselines used for comparison with \sysname.

\subsection{Datasets and Data Sources} \label{subsec:data-source}

\noindent\textbf{Benchmark Grids.}  
We compare \sysname against the state-of-the-art on the 13 grids used in prior studies.  
\emph{The US grids} are California~(CISO), Pennsylvania~(PJM), Texas~(ERCOT), New England~(ISO-NE), Washington~(BPAT), Florida~(FPL), and New York~(NYISO).  
\emph{The European grids} are Sweden~(SE), Germany~(DE), Poland~(PL), Spain~(ES), and the Netherlands~(NL).  
\emph{The Australian grid} is Queensland~(AUS-QLD).  
Data come from the CarbonCast repository~\cite{maji2022carboncast} and include two years of hourly carbon intensity measurements from 2020 to 2021.  
Following prior work, we evaluate forecasting accuracy on the final six months of 2021 for fair comparison.

\noindent\textbf{Global Carbon Intensity Dataset.}  
Electricity Maps recently released hourly carbon intensity data for 350 grids worldwide~\cite{emaps-datasets}. 
For global evaluation, we use data from 214 of these grids covering most major geographical areas, with the number of grids per area listed in Table~\ref{tab:worldwide-performance}.  
Each grid provides four years of hourly data from 2021 to 2024, split into training and testing sets with a 70–30 ratio.

\noindent\textbf{Uncertainty Quantification Dataset.}  
We evaluate uncertainty quantification on the same 13 benchmark grids using the two-year datasets from CarbonCast. We provide prediction intervals for the last 6 months in each dataset.

\noindent\textbf{Imputation Dataset.}  
We use 5 minute carbon intensity data from Electricity Maps~\cite{emaps-datasets} covering 50 grids, 30 in the United States and 20 in Europe.  
Each dataset spans 2021 to 2024 and is split 70 to 30 for training and testing.  
To simulate missing data, 12.5\% to 75\% of each time series is randomly masked for reconstruction.

\subsection{Performance Metrics}
\label{sec:metrics}

\noindent\textbf{Forecasting.}  
We use MAPE to assess forecasting accuracy.  
Lower values indicate better performance.  
In addition to the average MAPE over the test period, we report the 90th-percentile MAPE to capture tail forecasting errors of the models.

\noindent\textbf{Uncertainty Quantification.} We assess the quality of prediction intervals using \emph{Coverage} and \emph{Normalized Interval Width (NIW)}.

\emph{Coverage} measures the percentage of times the ground truth ($y_t$) falls within the prediction intervals ($[L_t, U_t]$):
\begin{equation}
\text{Coverage (\%)} = \frac{100}{T} \sum_{t=1}^{T} \mathbf{I} \left[ y_t \in [L_t, U_t] \right],
\end{equation}
where \( \mathbf{I}[\cdot] \) denotes the indicator function.

\emph{NIW} at each time step is the interval width relative to the ground truth: $\text{NIW}_t = \frac{U_t - L_t}{y_t}$.  
Lower NIW indicates a narrower interval.

\noindent\textbf{Imputation.}  
We use Root Mean Square Error (RMSE) on normalized time series to assess imputation accuracy.

\begin{table*}[t!]
\centering
\begin{threeparttable}
\resizebox{\textwidth}{!}{
\begin{tabular}{
    l
    cc|  
    cc|  
    cc|  
    cc|  
    cc|  
    cc|  
    cc|  
    cc|  
    cc  
}
\toprule

\text{} & 
\multicolumn{8}{c|}{\textbf{Baselines}} & 
\multicolumn{10}{c}{\textbf{\sysname Zero-Shot Mode}} \\

\cmidrule(lr){2-9} 
\cmidrule(lr){10-19}

\text{} & 
\multicolumn{2}{c}{\text{EWMA~\cite{shenoy2024home}}} & 
\multicolumn{2}{c}{\text{AutoARIMA~\cite{seabold2010statsmodels}}} & 
\multicolumn{2}{c}{\text{CarbonCast~\cite{maji2022carboncast}}\tnote{1}} & 
\multicolumn{2}{c|}{\text{EnsembleCI~\cite{yan2025ensembleci}}\tnote{1}} & 
\multicolumn{2}{c}{\text{MOMENT~\cite{goswami_moment_2024}}} & 
\multicolumn{2}{c}{\text{Chronos~\cite{ansari2024chronos}}} & 
\multicolumn{2}{c}{\text{TimesFM~\cite{das_decoder-only_2024}}} &
\multicolumn{2}{c}{\text{Sundial~\cite{liu2025sundial}}} & 
\multicolumn{2}{c}{\text{Time-MoE~\cite{shi2024time}}} \\
\cmidrule(lr){2-3} 
\cmidrule(lr){4-5} 
\cmidrule(lr){6-7} 
\cmidrule(lr){8-9} 
\cmidrule(lr){10-11} 
\cmidrule(lr){12-13} 
\cmidrule(lr){14-15} 
\cmidrule(lr){16-17} 
\cmidrule(lr){18-19} 

& Mean & 90th &
  Mean & 90th & 
  Mean & 90th & 
  Mean & 90th & 
  Mean & 90th & 
  Mean & 90th & 
  Mean & 90th & 
  Mean & 90th & 
  Mean & 90th 
  \\

\midrule

CISO & 12.93 & 28.15 
& 28.83 & 56.01 
& 11.45 & 22.21 
& \textbf{9.17} & 18.38 
& 23.38 & 48.49 
& 24.80 & 50.04 
& 24.31 & 48.41 
& 10.79 & \textbf{17.25}
& 11.84 & 24.79 \\

PJM & 6.18 & 13.01 
& 10.24 & 17.82 
& 5.29 & 8.74
& \textbf{3.41} & \textbf{7.17}
& 6.97 & 14.73 
& 9.40 & 17.83 
& 9.36 & 17.80 
& 4.68 & 7.98
& 4.56 & 9.86 \\

ERCOT & 17.33 & 37.45 
& 29.55 & 53.21 
& 11.14 & \textbf{21.44}
& \textbf{10.61} & 24.69 
& 17.14 & 37.15 
& 18.59 & 33.62 
& 18.40 & 32.69
& 14.64 & 25.46
& 15.45 & 35.82 \\

ISO-NE & 7.40 & 15.91 
& 11.04 & 20.67 
& 6.41 & \textbf{10.96}
& \textbf{5.74} & 12.98 
& 8.35 & 17.54 
& 12.20 & 22.17 
& 12.27 & 22.24 
& 6.89 & 11.52
& 6.79 & 14.99 \\

BPAT & 10.08 & 20.90 
& 19.33 & 32.79 
& 11.22 & 18.23
& 10.95 & 22.42 
& 12.47 & 25.53 
& 24.17 & 42.16 
& 24.32 & 42.30 
& \textbf{9.92} & \textbf{14.24}
& 10.01 & 20.43 \\

FPL & 3.28 & 7.35 
& 6.66 & 12.35
& 3.15 & \textbf{5.24} 
& \textbf{2.78} & 6.02 
& 4.96 & 9.91
& 6.93 & 13.72 
& 6.85 & 13.64 
& 2.86 & 5.31
& 2.85 & 6.39 \\

NYISO & 8.81 & 18.08
& 13.83 & 25.28 
& 9.09 & 28.09 
& 8.75 & 30.85
& 9.91 & 21.01 
& 17.45 & 35.01 
& 17.64 & 35.33  
& 8.59 & 16.19
& \textbf{7.86} & \textbf{15.97} \\

SE & 7.43 & 16.46 
& 14.09 & 26.32 
& 5.78 & \textbf{9.47}
& \textbf{5.42} & 11.63 
& 7.07 & 15.84 
& 8.83 & 17.71 
& 8.74 & 17.69 
& 6.22 & 10.37
& 6.37 & 13.62 \\

DE & 26.25 & 57.89 
& 42.39 & 95.86 
& \textbf{11.72} & \textbf{20.90} 
& 12.09 & 26.82 
& 27.05 & 58.59 
& 26.20 & 41.66 
& 25.71 & 40.67
& 21.06 & 40.59
& 20.46 & 43.23 \\

PL & 8.43 & 19.27 
& 16.26 & 33.22 
& 4.37 & \textbf{7.49}
& \textbf{4.23} & 9.34 
& 8.72 & 19.65 
& 7.79 & 14.78 
& 7.62 & 13.68 
& 6.48 & 10.91
& 6.42 & 13.26 \\

ES & 25.95 & 52.46 
& 34.69 & 69.24 
& \textbf{16.65} & \textbf{30.07} 
& 16.96 & 35.39 
& 25.53 & 53.99
& 23.75 & 43.53 
& 23.54 & 42.98 
& 20.42 & 34.99
& 19.68 & 39.73 \\

NL & 11.30 & 24.74 
& 22.43 & 40.87 
& 8.25 & \textbf{13.43}
& \textbf{6.96} & 13.72 
& 10.49 & 23.42 
& 10.48 & 18.12 
& 10.36 & 17.92
& 8.69 & 15.51
& 8.78 & 17.69 \\

AUS-QLD & 4.04 & 9.67 
& 23.79 & 46.67 
& 4.46 & 6.72
& 3.72 & 8.99 
& 15.47 & 33.56 
& 15.76 & 39.42 
& 15.46 & 39.30 
& \textbf{3.48} & \textbf{4.69}
& 3.56 & 8.49 \\

\bottomrule

\textbf{Average} & 11.49 & 24.72 
& 21.00 & 40.80 
& 8.38 & \textbf{15.61}
& \textbf{7.75} & 17.57 
& 13.63 & 29.18
& 15.87 & 29.98
& 15.73 & 29.59
& 9.59 & 16.54
& 9.59 & 20.25 \\

\bottomrule
\end{tabular}
}
\begin{tablenotes}
\small
\item[(1)] CarbonCast and EnsembleCI require energy mix information and train grid-specific models to forecast carbon intensity.

\end{tablenotes}
\end{threeparttable}
\caption{\emph{4-day MAPE comparison (best results in \textbf{bold}). TSFMs provide competitive performance in some grids. Sundial (with a 30-day lookback) achieves the highest accuracy in the zero-shot mode, having 9.59\% mean and 16.54\% tail forecasting MAPE, compared to state-of-the-art mean MAPE of 7.75\% (EnsembleCI) and tail MAPE of 15.61\% (CarbonCast).}}
\label{tab:sota-vs-tier1-zero-shot}
\end{table*}

\subsection{Baselines}
\noindent\textbf{Forecasting.} We compare \sysname with two state-of-the-art tools:  CarbonCast~\cite{maji2022carboncast} and EnsembleCI~\cite{yan2025ensembleci}.
While EnsembleCI performs better than CarbonCast on average, CarbonCast has better tail forecasting accuracy (see Table~\ref{tab:comparing-sota-with-our-work}). 
Hence, we include both tools as baselines. 
We also implement two moving average-based statistical baselines: EWMA~\cite{shenoy2024home} and AutoARIMA~\cite{seabold2010statsmodels}.

\vsnoindent 
We could not compare \sysname with services such as Electricity Maps~\cite{electricitymaps} and Watttime~\cite{watttime}, which also provide worldwide forecasts, because their forecasts are proprietary and not public.

\noindent\textbf{Uncertainty Quantification.} 
Li \etal~\cite{li2024uncertainty} provide prediction intervals for carbon intensity forecasts. While both our and their works provide the desired coverage, comparing NIW directly is non-trivial, as the interval width depends on the accuracy of the underlying point-value forecasts. Since Li \etal extends CarbonCast, which has a higher accuracy than \sysname, our NIW will potentially be higher. We keep this comparison as future work. 

Importantly, we note that \sysname's modular architecture enables seamless integration of any state-of-the-art methods that can provide narrower intervals with the available \sysname forecasts.

\noindent\textbf{Imputation.} We compare \sysname's zero-shot and fine-tuned imputation performance against three statistical baselines: (1) Naive interpolation, which fills a missing value using the same hour from the nearest available day (in the past or future), (2) Linear interpolation, and (3) Cubic spline interpolation. Linear and cubic spline interpolation are standard statistical baselines that perform well in estimating missing data in time series. The naive approach simulates a realistic operational heuristic often used by practitioners.

\subsection{Fine-Tuning Strategies}

During forecasting, we fine-tune the TSFMs for one epoch for both grid- and domain-specific strategies. During imputation, we employ both light-weight and full fine-tuning for five epochs. For imputation, we only opt for grid-specific fine-tuning.

\begin{table}[t]
\centering
\begin{threeparttable}
\resizebox{\columnwidth}{!}{
\begin{tabular}{
    l
    cc|  
    cc|  
    cc|  
    cc  
}
\toprule
& \multicolumn{8}{c}{\textbf{\sysname Grid-Specific Fine-Tuning}} \\
\cmidrule(lr){2-9}
& \multicolumn{2}{c}{\text{MOMENT~\cite{goswami_moment_2024}}} &
  \multicolumn{2}{c}{\text{Chronos~\cite{ansari2024chronos}}} &
  \multicolumn{2}{c}{\text{TimesFM~\cite{das_decoder-only_2024}}} &
  \multicolumn{2}{c}{\text{Time-MoE~\cite{shi2024time}}} \\
\cmidrule(lr){2-3}
\cmidrule(lr){4-5}
\cmidrule(lr){6-7}
\cmidrule(lr){8-9}
& Mean & 90th & Mean & 90th & Mean & 90th & Mean & 90th \\
\midrule
CISO     & 12.27 & 25.07 & 23.51 & 43.57 & 12.47 & 26.79 & 12.03 & 25.54 \\

PJM      &  5.24 & 10.84 & 9.50 & 17.60 &  5.21 & 11.46 & 5.20 & 11.32 \\

ERCOT    & 16.06 & 33.66 & 18.90 & 35.64 & 15.59 & 33.10 & 17.35 & 37.50 \\

ISO-NE   &  6.99 & 14.26 & 12.69 & 22.17 &  7.23 & 15.43 & 7.21 & 15.02 \\

BPAT     & 10.45 & 21.68 & 24.85 & 41.91 & 10.45 & 21.67 & 10.05 & 20.61 \\

FPL      &  3.43 &  7.00 & 7.98 & 14.23 &  3.03 &  6.44 & 3.30 & 7.14 \\

NYISO    &  7.95 & 15.77 & 16.50 & 38.02 & 10.14 & 24.35 & 8.39 & 17.24 \\

SE       &  6.59 & 14.85 & 10.31 & 18.53 &  6.19 & 14.12 & 7.07 & 15.31 \\

DE       & 22.07 & 41.45 & 26.27 & 42.64 & 22.21 & 47.46 & 23.48 & 53.72 \\

PL       &  6.92 & 13.74 & 8.35 & 15.73 &  7.01 & 15.21 & 7.55 & 17.29 \\

ES       & 20.85 & 42.12 & 24.99 & 44.14 & 22.84 & 46.84 & 23.45 & 45.49 \\

NL       &  9.38 & 18.16 & 11.47 & 18.63 &  8.96 & 19.59 & 9.88 & 20.45 \\

AUS-QLD  &  4.83 & 11.82 & 16.80 & 38.61 &  3.69 &  8.53 & 3.72 & 8.73 \\

\midrule
\textbf{Average}
         & 10.23 & 20.80 & 16.32 & 30.11 & 10.39 & 22.38 & 10.66 & 22.72 \\
\bottomrule
\end{tabular}
}
\caption{\emph{\sysname accuracy (MAPE \%) after grid-specific fine-tuning (Sundial is excluded from evaluation).}}
\label{tab:sota-vs-tier1-fine-tuned}
\end{threeparttable}
\end{table}

\begin{table}[t]
\centering
\begin{threeparttable}
\resizebox{0.71\columnwidth}{!}{
\begin{tabular}{l cc| cc}
\toprule
& \multicolumn{4}{c}{\textbf{\sysname Domain-Specific Fine-Tuning}} \\
\cmidrule(lr){2-5}
& \multicolumn{2}{c}{\text{Chronos~\cite{ansari2024chronos}}} &
  \multicolumn{2}{c}{\text{Time-MoE~\cite{shi2024time}}} \\
\cmidrule(lr){2-3}
\cmidrule(lr){4-5}
& Mean & 90th & Mean & 90th \\
\midrule
CISO     & 22.39 & 40.64 & 12.41 & 27.07 \\
         
PJM      &  9.04 & 17.28 &  4.97 &  10.87 \\

ERCOT    & 19.11 & 38.57 &  16.57 & 35.66 \\
         
ISO-NE   & 11.70 & 21.45 &  7.09 & 15.50 \\
         
BPAT     & 24.61 & 41.67 & 9.83 & 20.29 \\
         
FPL      &  7.15 & 13.80 &  3.02 &  6.66 \\
         
NYISO    & 16.20 & 38.35 &  8.98 & 18.56 \\
         
SE       & 9.24 & 18.11 &  6.71 & 14.78 \\
         
DE       & 25.80 & 40.76 & 23.20 & 49.65 \\
         
PL       &  7.70 & 15.14 &  7.25 & 15.58 \\
         
ES       & 24.15 & 43.84 & 23.42 & 45.80 \\
         
NL       & 10.41 & 18.16 
         &  9.50 & 19.79 \\
         
AUS-QLD  & 16.23 & 39.71 &  3.98 &  9.30 \\
         
\midrule
\textbf{Average}
         & 15.67 & 29.80 &  10.53 & 22.27 \\
\bottomrule
\end{tabular}
}
\caption{\emph{\sysname accuracy (MAPE \%) after domain-specific fine-tuning (only Chronos and Time-MoE are evaluated).}}
\label{tab:sota-vs-tier1-global-finetune}
\end{threeparttable}
\end{table}

\section{Forecasting Task Evaluation}
\label{sec_evaluation}

We evaluate \sysname across multiple dimensions.  
First, we compare its accuracy on 13 benchmark grids in both zero-shot and fine-tuned modes and show that \sysname matches the tail performance of the current state-of-the-art. Next, we show that its forecasts remain reliable for horizons up to 21 days. We then demonstrate that \sysname can provide accurate forecasts globally, by evaluating across 214 grids worldwide. Finally, we assess uncertainty quantification, confirming well-calibrated prediction intervals.  

\subsection{Comparison against Baselines}
\label{sec:sota-comparison}

Tables~\ref{tab:sota-vs-tier1-zero-shot}, \ref{tab:sota-vs-tier1-fine-tuned}, and \ref{tab:sota-vs-tier1-global-finetune} present detailed forecasting accuracy of \sysname across 13 grids in both zero-shot and fine-tuned modes.

\subsubsection{\textbf{Zero-Shot TSFM Performance.}}

Statistical models provide only moderate accuracy. 
EWMA achieves an average mean MAPE of 11.49\%, outperforming AutoARIMA at 21.00\%, and performs well on stable grids such as FPL, where it records 3.28\% mean MAPE but deteriorates on volatile grids such as DE, where it reaches 26.25\%.  
CarbonCast and EnsembleCI remain the strongest domain-specific models, leveraging energy mix features and grid-specific architecture and training to achieve mean and 90th-percentile MAPE of 8.38\% and 15.61\%, and 7.75\% and 17.57\%, respectively.

\vsnoindent
Among the pre-trained TSFMs, Sundial with a 30-day lookback leads with an average mean and 90th-percentile MAPE of 9.59\% and 16.54\%. Sundial is strong in stable grids such as PJM and FPL, where it reaches single-digit MAPE, and it maintains robustness in higher variance grids such as DE and ES, even outperforming EnsembleCI on tail forecasting accuracy.
Time-MoE delivers similar zero-shot accuracy as Sundial, having a similar mean MAPE (9.59\%) but slightly higher tail errors (20.25\%).
MOMENT follows with 13.63\% mean and 29.18\% 90th-percentile MAPE, benefiting from strong performance on North American grids but dropping sharply on European ones.  
Chronos records 15.87\% mean and 29.98\% 90th-percentile MAPE, and TimesFM records 15.73\% mean and 29.59\% 90th-percentile MAPE, reflecting weaker adaptation to the heterogeneous temporal dynamics of carbon intensity data.

We also evaluate other variants of Sundial and Time-MoE to assess the effects of model size on accuracy. Accuracy increases with model size for most grids, showing that larger models can typically capture the patterns better. However, in some grids, smaller models perform better --- especially for Time-MoE. The results are available in Appendix~\ref{appendix:zero-shot-variants}, Table~\ref{tab:zero-shot-variants}. These results underscore the need for more in-depth analysis of model size, architecture, and performance.

\vsnoindent
\textbf{Summary.} \sysname can deliver competitive accuracy across diverse grids using \emph{just one model}, only historical carbon intensity data, and no grid-specific modifications. This highlights the strength and versatility of \sysname, making it a strong candidate for global carbon intensity forecasting. Among the TSFMs, Sundial achieves the highest accuracy, providing robust tail performance that \emph{outperforms EnsembleCI}.

\begin{table}[t!]
\centering
\resizebox{0.9\columnwidth}{!}{
\begin{tabular}{l|
                                c|
                                cc|
                                cc}
\toprule

\text{} & 
\multicolumn{1}{c|}{\text{}} & 
\multicolumn{2}{c|}{\textbf{ \sysname}} &
\multicolumn{2}{c}{\textbf{ \sysname}} \\

\text{} & 
\multicolumn{1}{c|}{\text{}} & 
\multicolumn{2}{c|}{\textbf{(w/ Sundial) }} &
\multicolumn{2}{c}{\textbf{(w/ Time-MoE) }} \\


 
\cmidrule(lr){3-4}
\cmidrule(lr){5-6}

Area & \# of grids &
Mean & 90th &
Mean & 90th \\

\midrule

Africa & 28 & 4.72 & 7.96 & 4.76 & 10.79  \\

Asia & 39 & 3.87 & 6.91 & 4.35 & 9.28 \\

Central America & 9 & 16.78 & 33.05 & 16.39 & 39.18 \\

Europe & 39 & 21.44 & 39.10 & 22.92 & 52.33\\

North America & 73 & 16.10 & 28.86 & 16.90 & 35.36\\

Oceania & 11 & 73.57 & 153.59 & 82.02 & 244.53 \\

South America & 15 & 8.64 & 15.07 & 10.70 & 25.45\\

\bottomrule

\textbf{Total/Average} & 214 & 15.82 & 29.61 & 17.01 & 33.37\\
\bottomrule
\end{tabular}
}
\caption{\emph{Zero-shot forecasting accuracy (MAPE \%) of \sysname aggregated over 214 grids worldwide. With Sundial, \sysname has an average MAPE of 15.82\%.}}
\label{tab:worldwide-performance}
\end{table}

\subsubsection{\textbf{Grid-Specific Fine-Tuning}}
To enhance performance beyond the zero-shot setting, we evaluate grid-specific fine-tuning of the TSFMs. 
Table~\ref{tab:sota-vs-tier1-fine-tuned} reports the 4-day MAPE when each model is trained on the historical carbon intensity of a single grid and tested only on that grid’s hold-out data. 
Sundial is excluded because it currently lacks a fine-tuning interface.

\vsnoindent
Fine-tuning yields clear gains for some models. 
MOMENT’s average mean MAPE drops from 13.63\% to 10.23\%, and its 90th-percentile error falls from 29.18\% to 20.80\%. 
TimesFM improves in a similar way, with mean MAPE decreasing from 15.73\% to 10.39\% and the 90th-percentile from 29.59\% to 22.38\%. 
These improvements arise because their architectures can adapt with only the final linear layer retrained on grid-specific data, allowing the shared representations to specialize without overfitting.

\vsnoindent
In contrast, Chronos and Time-MoE show little benefit and even slight degradation on average. 
Chronos rises from 15.87\% to 16.32\% mean MAPE, while Time-MoE moves from 9.59\% to 10.66\%. 
Chronos relies on a fixed, discrete vocabulary to encode time-series values, which may be difficult to refine with limited regional data, leading to overfitting or unstable token representations. 
Time-MoE employs a mixture-of-experts routing mechanism that likely requires larger-scale or more carefully balanced updates; small per-grid datasets may disrupt expert activation and reduce overall accuracy.

\vsnoindent\textbf{Summary.}
Grid-specific fine-tuning enables some TSFMs, such as  MOMENT and TimesFM, to close the accuracy gap with CarbonCast and EnsembleCI. 
Interestingly, zero-shot Sundial still outperforms other TSFMs even after grid-specific fine-tuning.

\subsubsection{\textbf{Domain-Specific Fine-Tuning}}

We train single instances of Chronos and Time-MoE on the combined training data from 214 grids and evaluate their accuracy on the 13 benchmark grids. Table~\ref{tab:sota-vs-tier1-global-finetune} presents the mean and tail MAPE of both models. 

Chronos shows marginal improvements on average, with mean MAPE moving from 15.87\% in the zero-shot mode to 15.67\%. However, the tail MAPE shifts from 29.18\% to 29.80\%. The larger, more heterogeneous dataset offers small gains on average but introduces variability that limits gains in the tail. On the other hand, although Time-MoE's accuracy improves over its grid-specific version, it still falls behind its zero-shot accuracy. The mean (resp. tail) MAPE is 10.53\% (resp. 22.27\%), which is lower than 10.66\% (resp. 22.72\%) achieved with grid-specific fine-tuning, but is higher than 9.59\% (resp. 20.25\%) achieved in the zero-shot mode. 

One reason for such behaviour can be that since we are doing full fine-tuning and the size of the new dataset is negligible compared to the pre-training dataset, the model weights are updated insignificantly. 
Thus, although our preliminary results show promise with this approach, more careful analysis and design are required for conclusive results.

\vsnoindent
\textbf{Summary.}  
Domain-specific fine-tuning provides marginal improvements over grid-specific fine-tuning. However, more detailed analysis is required to fully realize the potential of this approach.

\begin{figure}[t!]
    \centering
    \includegraphics[width=0.8\linewidth]{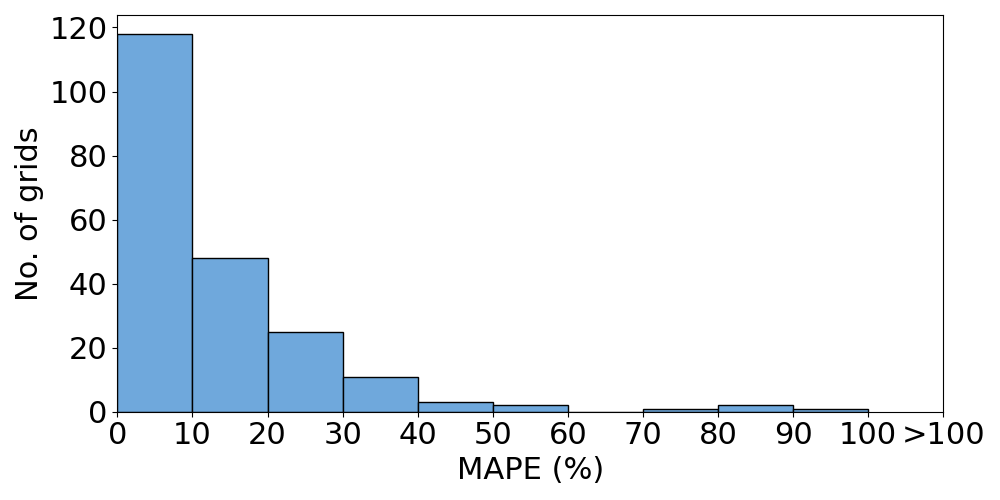}
    \caption{\emph{MAPE distribution of \sysname across 214 grids using Sundial as the TSFM. 166 grids have < 20\% MAPE on average.}}
    \label{fig:continent-histogram}
    \Description[]{}
    \vspace{-0.1cm}
\end{figure}

\subsection{Worldwide Evaluation}
Next, we evaluate \sysname globally across 214 grids spanning every continent except Antarctica. We employ the most generalized setting, where one model forecasts all the grids in zero-shot mode. Since Sundial and Time-MoE are the best-performing models in the zero-shot mode across the 13 benchmark grids, we evaluate the accuracy by plugging both these TSFMs into our task module.

\vsnoindent
Table~\ref{tab:worldwide-performance} summarizes the number of grids evaluated in each geographical area and the forecasting accuracy of both TSFMs in each area. We report both the average forecasting accuracy (mean MAPE) and the tail forecasting accuracy (90th percentile MAPE). Sundial delivers strong performance globally, achieving an average MAPE of 15.82\% and a tail forecasting accuracy of 29.61\% across the 214 grids. Time-MoE follows Sundial performance closely, with an average accuracy of 17.01\% and a tail accuracy of 33.37\%. 

\vsnoindent
Figure~\ref{fig:continent-histogram} and Table~\ref{tab:worldwide-performance} further break down the overall forecasting accuracy of \sysname using Sundial as the TSFM by showing the worldwide error distribution. Most grids have a low forecasting error, with 166 out of 214 grids having a mean MAPE below 20\%. Grids in Asia and Africa, which have a high percentage of fossil fuels and thus more stable carbon intensity profiles, have highly accurate forecasts, with an average MAPE of less than 5\%. On the other hand, grids in the Americas and Europe have more renewables, leading to greater temporal volatility and increasing the forecasting errors. However, \sysname still achieves high accuracy, with 96-hour forecasting errors being below 20\% in most of the grids. The only exception is Oceania, where \sysname has a very high forecasting error. This high error rate is primarily due to three grids in Tasmania and its neighboring islands, where frequent large swings in renewable share make the grids highly volatile and, hence, forecasting significantly difficult. If these three grids are excluded, the mean and tail MAPE in Oceania decrease to 20.69\% and 36.20\%, respectively, similar to other geographical areas. Consequently, the global mean and tail MAPE across 211 grids also decrease to 12.99\% and 23.40\%, respectively, further underscoring \sysname's strength.

\vsnoindent
We observe similar accuracy characteristics and regional patterns when using Time-MoE as the TSFM in \sysname. 

\vsnoindent
\textbf{Summary.} \sysname delivers strong global zero-shot forecasting performance using only historical carbon intensity data: 77.5\% of the evaluated 214 grids achieve mean MAPE under 20\%. Notably, in more than 35\% of these grids, electricity mix data is not available, making prior works infeasible and \sysname the \emph{only feasible option}. Even if such data were available, prior works would have required hundreds of models to provide global coverage, while \sysname can provide that using only \emph{one generalized model}. Thus, \sysname can be used as a tool globally without additional modifications, making it highly practical for computational decarbonization. 

\begin{table}[t!]
\centering
\resizebox{0.9\linewidth}{!}{
\begin{tabular}{>{\raggedright\arraybackslash}p{4.3em} 
                >{\centering\arraybackslash}p{3em} 
                >{\centering\arraybackslash}p{3em} 
                >{\centering\arraybackslash}p{4em} 
                >{\centering\arraybackslash}p{3.2em} 
                >{\centering\arraybackslash}p{4em}}
\toprule
\textbf{} & \textbf{D1} & \textbf{D4} & \textbf{Drop\textsubscript{4}} & \textbf{D21} & \textbf{Drop\textsubscript{21}}\\
\midrule

CISO     & 8.03 & 10.96 & 2.93 & 12.38 & 4.35 \\
PJM      & 3.12 & 4.76  & 1.64 & 5.89  & 2.77 \\
ERCOT    & 11.63 & 15.00 & 3.37 & 15.84 & 4.21 \\
ISO-NE   & 4.91 & 6.99  & 2.08 & 9.90  & 4.99 \\
BPAT     & 6.93 & 9.43  & 2.50 & 12.97 & 6.04 \\
FPL      & 2.09 & 2.97  & 0.88 & 4.19  & 2.10 \\
NYISO    & 5.18 & 8.96  & 3.78 & 15.46 & 10.28 \\
SE       & 4.26 & 6.28  & 2.02 & 7.91  & 3.65 \\
DE       & 14.83 & 21.15 & 6.32 & 24.79 & 9.96 \\
PL       & 4.31 & 6.58  & 2.27 & 7.34  & 3.03 \\
ES       & 11.84 & 21.47 & 9.63 & 23.87 & 12.03 \\
NL       & 6.06 & 8.88  & 2.82 & 9.98  & 3.92 \\
AUS-QLD  & 2.81 & 3.59  & 0.78 & 4.05  & 1.24 \\
\midrule
\textbf{Average}  & 6.62 & 9.77  & 3.16 & 11.89 & 5.27 \\



\bottomrule
\end{tabular}
}
\caption{\emph{Extended forecasting accuracy~(MAPE) and degradation (absolute difference in MAPE) compared to D1 forecasts. Sundial with a 15-day lookback provides the best results.}}
\label{tab:long-horizon-forecast-degradtion}
\end{table}

\subsection{Extended Forecasting}

TSFMs have shown strong long-horizon forecasting across various domains. We therefore hypothesize that TSFMs can forecast carbon intensity accurately for up to several weeks. In contrast, existing carbon forecasting tools are typically designed and evaluated for horizons up to 96 hours. Their accuracy beyond this horizon is expected to decrease substantially due to their limited model capacity. Further, these tools rely on auxiliary information such as weather forecasts, which are only available for up to 16 days in many regions~\cite{weather-forecasts}. This creates a practical bottleneck for existing tools to provide extended carbon intensity forecasts. 

\vsnoindent
In this section, we assess \sysname's extended forecasting capabilities by evaluating its zero-shot accuracy for up to 3 weeks (21 days). Since Sundial achieves the highest accuracy while providing zero-shot 4-day forecasts, we use Sundial as the TSFM for this experiment. We vary the lookback window from 1 to 30 days in the past, and extend the forecasting horizon from 1 to 21 days. While a 30-day lookback achieves the highest accuracy for 4-day forecasts, a 15-day lookback yields the lowest accuracy degradation for 21-day forecasts across the grids. We posit that this is because while shorter lookbacks cannot capture the longer temporal patterns, longer lookbacks may incorporate noise or infrequent events when forecasting longer horizons. A 15-day lookback achieves the best balance, capturing the patterns without amplifying noise. More detailed ablations of lookback length and forecast horizon are provided in Appendix~\ref{appendix:long_forecasting}, Table~\ref{tab:long-horizon-forecast-degradtion_appendix}.

\vsnoindent
Table~\ref{tab:long-horizon-forecast-degradtion} summarizes the results across the 13 benchmark grids. Sundial's has a high day-ahead accuracy, with an average MAPE of 6.62\%. The error growth over time is modest: the 4-day horizon averages 9.77\% MAPE, only 3.16\% points above day-ahead, and the 21-day horizon averages 11.89\%, just 5.27 points higher. This limited increase underscores \sysname’s ability to maintain reliable performance even for longer forecasts.

\vsnoindent
The degradation depends on the carbon intensity volatility. Stable grids (\eg PJM, FPL) show exceptional accuracy, having less than 3\% degradation and MAPE below 6\% even at 21 days. More volatile grids (\eg Germany (DE), Spain (ES)) show larger degradation: DE rising from 14.83\% to 24.79\% and ES from 11.84\% to 23.87\%. Still, the degradation remains manageable over three weeks. Notably, most of the degradation occurs in the first four days; then the degradation rate decreases. For example, in DE, accuracy decreases from 14.83\% on day one to 21.15\% by day 4. However, the accuracy only degrades by 3.64\% over the remaining forecasting horizon.

\begin{table}[t]
\centering
\begin{threeparttable}
\resizebox{0.95\linewidth}{!}{
\begin{tabular}{
    l|
    c|  
    ccccc|  
    c  
}
\toprule

\text{} & 
\multicolumn{1}{c}{} & 
\multicolumn{5}{c}{\textbf{Coverage (\%)}} & 
\multicolumn{1}{c}{} \\

\cmidrule(lr){3-7} 

& \textbf{MAPE (\%)} &
D1 & D2 & D3 & D4 & Overall & \textbf{NIW (\%)}
  \\

\midrule

CISO     & 10.79 & 95.6 & 95.9 & 96.7 & 96.9 & 96.3 & 68.5 \\
PJM      & 4.68  & 96.4 & 96.1 & 96.6 & 96.4 & 96.4 & 29.0 \\
ERCO     & 14.64 & 95.6 & 94.7 & 95.1 & 94.6 & 95.0 & 67.7 \\
ISO-NE     & 6.89  & 97.6 & 97.3 & 96.5 & 96.0 & 96.9 & 41.6 \\
BPAT     & 9.92  & 96.9 & 95.9 & 95.8 & 95.7 & 96.1 & 53.1 \\
FPL      & 2.86  & 96.0 & 95.3 & 95.2 & 94.5 & 95.3 & 21.6 \\
NYISO    & 8.59  & 97.4 & 96.7 & 95.7 & 94.7 & 96.1 & 70.1 \\
SE       & 6.22  & 96.5 & 95.6 & 94.7 & 94.3 & 95.3 & 36.0 \\
DE       & 21.06 & 96.7 & 97.7 & 97.2 & 95.3 & 96.7 & 103.6 \\
PL       & 6.48  & 97.5 & 96.6 & 96.4 & 96.0 & 96.6 & 36.9 \\
ES       & 20.42 & 96.5 & 94.3 & 94.6 & 93.7 & 94.8 & 96.3 \\
NL       & 8.69  & 97.4 & 97.4 & 97.3 & 96.9 & 97.3 & 49.6 \\
AUS-QLD  & 3.48  & 97.6 & 97.5 & 98.8 & 98.5 & 98.1 & 30.8 \\

\bottomrule

\textbf{Average} 

& 9.59 & 96.7 & 96.2 & 96.2 & 95.6 & 96.2 & 54.2

\\

\bottomrule
\end{tabular}
}
\end{threeparttable}
\caption{\emph{Uncertainty quantification: \sysname provides more than the desired 95\% coverage with moderately wide intervals.}}
\label{tab:uncertainty-coverage-niw}
\end{table}

\vsnoindent
\textbf{Summary.} \sysname delivers strong zero-shot accuracy with \emph{minimal extended-horizon degradation} (5.27\% accuracy drop across 21 days). 
Given an appropriate historical context, \sysname provides reliable forecasts for applications that require horizons well beyond conventional day-ahead planning.

\subsection{Uncertainty Quantification}

\begin{figure}[t!]
    \centering
    \includegraphics[width=0.85\linewidth]{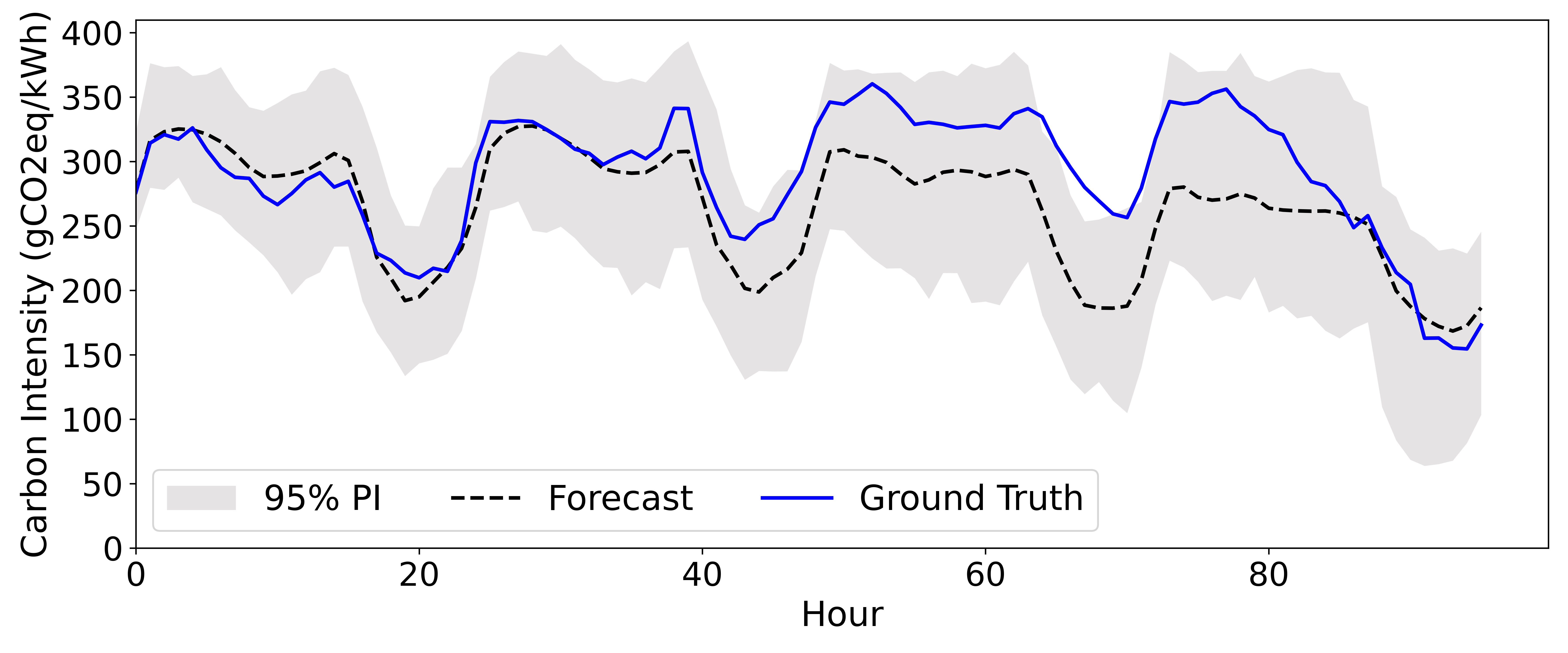}
    \caption{\emph{\sysname forecasts and prediction intervals during a 96-hour forecasting period in the CISO grid. 
    The intervals provide desired coverage around the ground truth.}
    }
    \label{fig:prediction-intervals}
    \Description[]{}   
\end{figure}

We now evaluate how \sysname quantifies uncertainty associated with the carbon intensity forecasts using the lightweight conformal layer. Figure~\ref{fig:prediction-intervals} shows the 96-hour prediction intervals produced by \sysname for the California (CISO) grid. The prediction intervals are constructed from the most recent historical residuals, which ensures that the intervals closely follow the ground truth trajectory and provide the desired coverage (95\%). This demonstrates the robustness of the conformal layer and the reliability that \sysname offers to downstream applications that may make optimization decisions based on these forecasts and intervals.

\vsnoindent
Table~\ref{tab:uncertainty-coverage-niw} reports the empirical Coverage and Normalized Interval Width (NIW) results across the 13 benchmark grids. \sysname provides more than 95\% coverage --- both daywise and over the whole 96-hour horizon --- when averaged across the grids. At an individual grid level, \sysname provides an average coverage above 95\% in all grids except Spain (ES), where the coverage dips slightly below 95\% due to increased uncertainty in forecasting beyond the 24-hour horizon, as evident from the Table. 

\vsnoindent
The NIW varies from 21.6\% in Florida (FPL) to 103.6\% in Germany (DE), with an average of 54.2\% across the grids. In general, NIW has a positive correlation with the forecast errors: grids with less accurate point forecasts tend to have wider prediction intervals. This is expected behaviour, as greater uncertainty and hence residual variance require wider intervals to maintain coverage guarantees.

\vsnoindent
An average NIW of 54.2 signifies that the interval width is slightly more than half of the magnitude of the ground truth values on average. We note that the prediction intervals are considerably wide in certain grids (e.g., Germany). Decreasing the intervals while maintaining coverage may require more advanced conformal prediction techniques~\cite{xu2023sequential} than our current approach. Notably, our modular design allows seamless integration of such techniques with minimal changes, and we keep integrating advanced conformal prediction techniques as future work.

\vsnoindent
\textbf{Summary.} \sysname's conformal layer effectively quantifies forecast uncertainty by producing adaptive, well-calibrated prediction intervals that provide the desired 95\% coverage while maintaining 54.2\% NIW. These results enable \sysname to be a practical tool with reliable and robust forecasts for downstream applications that require carbon-aware decision-making under uncertainty.

\begin{figure}[t]
\centering
\begin{subfigure}[t]{0.49\columnwidth}
  \centering
  \includegraphics[width=\linewidth]{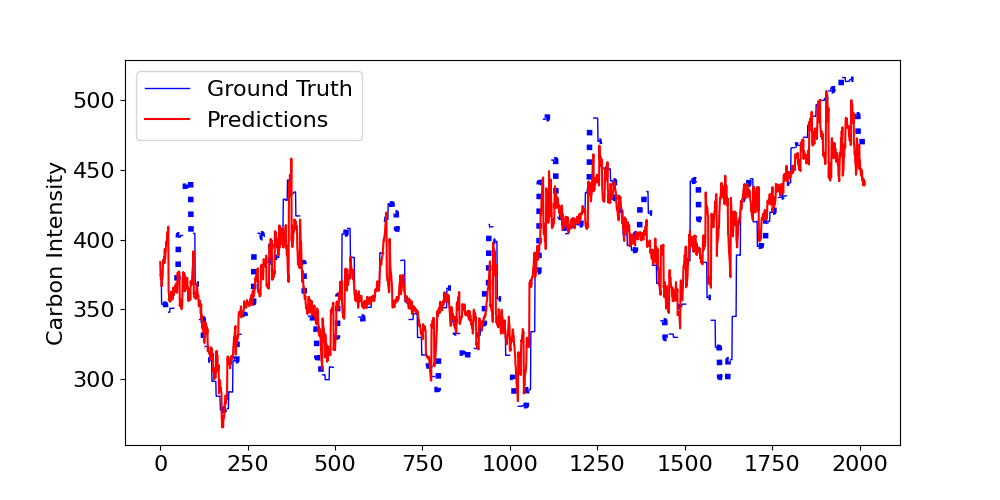}
  \caption{\emph{Zero-shot (ERCOT).}}
  \label{fig:erco-imputation-mask500-zs}
\end{subfigure}
\hfill
\begin{subfigure}[t]{0.5\columnwidth}
  \centering
  \includegraphics[width=\linewidth]{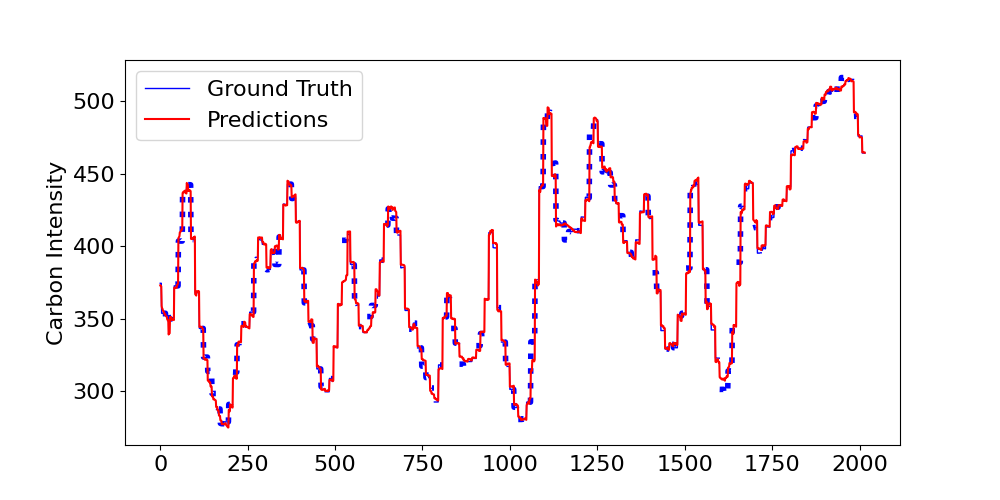}
  \caption{\emph{Fine-tuned (ERCOT).}}
  \label{fig:erco-imputation-mask500-ft}
\end{subfigure}
\caption{\emph{\sysname provides good estimates of missing data.}}
\label{fig:erco-imputation-comparison}
\end{figure}

\section{Imputation Task Evaluation}
\label{sec:imputation}

Finally, we evaluate carbon intensity imputation by masking random fixed-length patches of the data to simulate scenarios where some of the data is missing. A masking percentage of 50\% means that we simulate a scenario where half of the data is missing. We vary this masking percentage from 12.5\% to 75\% during our experiments, and evaluate for 50 grids (30 in the US and 20 in Europe).

\vsnoindent
Figure~\ref{fig:erco-imputation-comparison} visualizes \sysname's imputation performance for the Texas grid (ERCOT), both in zero-shot and fine-tuned modes. The blue segments indicate the ground truth, with gaps denoting missing data. \sysname estimates follow the ground truth in both modes, with fine-tuned estimates being more accurate. 

\vsnoindent
Table~\ref{tab:imputation-rmse} details the imputation accuracy of \sysname in different modes and compares \sysname against three statistical baselines. Since we average the RMSE across multiple grids that have different carbon intensity magnitudes, we first normalize the values for fair comparison. The naive and cubic spline baselines perform modestly, with RMSE between 0.83 and 0.94 across the masking percentages. Zero-shot \sysname is substantially better, having 0.18 RMSE (resp. 0.52) with 12.5\% (resp. 75\%) missing data. With lightweight fine-tuning, the accuracy improves further: RMSE reduces to 0.41 for 75\% missing data. Interestingly, we observed that the linear interpolation method performs better in terms of RMSE, due to occasional large estimate deviation in \sysname. However, the linear method connects endpoints with straight lines and can miss temporal patterns. Thus, in pattern-sensitive applications, lightly fine-tuned \sysname may be preferable over linear interpolation.

\vsnoindent
To further improve the accuracy, we also fully fine-tune \sysname per grid. This mode outperforms all baselines, achieving an RMSE of 0.07 (resp. 0.24) with a 12.5\% (resp. 75\%) masking percentage, a 1.2---3.9$\times$ improvement. We note that full fine-tuning yields grid-specific models, and hence, may be challenging to deploy at scale; however, we report the values so that users can decide the mode per their requirements.


\vsnoindent
\textbf{Summary.} \sysname provides moderate to high imputation accuracy for missing data. While linear interpolation can provide better RMSE than lightweight fine-tuning, \sysname preserves the patterns better on average. Fully fine-tuned version achieves the best results overall, outperforming all baselines by 1.2---3.9$\times$.

\begin{table}[t]
\centering
\resizebox{0.85\linewidth}{!}{%
\begin{tabular}{
  >{\raggedright\arraybackslash}p{6em}
  >{\centering\arraybackslash}p{2em}
  >{\centering\arraybackslash}p{2em}
  >{\centering\arraybackslash}p{2em}
  >{\centering\arraybackslash}p{2em}
  >{\centering\arraybackslash}p{2em}
  >{\centering\arraybackslash}p{2em}}
\toprule
\textbf{} & \textbf{12.5\%} & \textbf{25\%} & \textbf{37.5\%} & \textbf{50\%} & \textbf{62.5\%} & \textbf{75\%}\\
\midrule
\textbf{FT (full)}   & \textbf{0.07} & \textbf{0.08} & \textbf{0.11} & \textbf{0.14} & \textbf{0.18} & \textbf{0.24} \\
\textbf{FT (head)}   & 0.19 & 0.21 & 0.24 & 0.28 & 0.34 & 0.41 \\
\textbf{ZS}          & 0.18 & 0.23 & 0.30 & 0.36 & 0.44 & 0.52 \\
\textbf{Naive}       & 0.83 & 0.85 & 0.86 & 0.89 & 0.91 & 0.94 \\
\textbf{Cubic-Spline}& 0.86 & 0.86 & 0.86 & 0.86 & 0.87 & 0.87 \\
\textbf{Linear}      & 0.13 & 0.14 & 0.15 & 0.17 & 0.21 & 0.29 \\
\bottomrule
\end{tabular}
}
\caption{Normalized imputation RMSE for different methods at various masking percentages.}
\label{tab:imputation-rmse}
\end{table}

\section{Conclusion}
\label{sec_conclusion}

We present \sysname, an open-sourced tool that leverages the versatility and strong performance of TSFMs to perform carbon intensity forecasting and imputation globally.  
\sysname achieves strong zero-shot performance across 214 grids globally using only historical carbon data, with a mean MAPE of 15.62\%. It supports forecasting up to 21 days, provides well-calibrated prediction intervals with desired coverage, and performs accurate imputation, making it a practical and easy-to-use tool for carbon-aware applications.
We release code and datasets to foster future sustainability research.


\section*{Acknowledgments}
This work is supported in part by NSF grants CNS-2105494, CNS-2325956, 23091241, 2213636, 2211302, and 2211888, DOE grant DEEE0010143, and a grant from VMware. 
Mani Srivastava was also partially supported by the Mukund Padmanabhan Term Chair at UCLA.

\bibliographystyle{ACM-Reference-Format}
\bibliography{references}

\appendix

\begin{table*}[t]
\centering
\begin{tabularx}{\textwidth}{l l l X}
\toprule
\textbf{Function} & \textbf{Parameters} & \textbf{Returns} & \textbf{Description} \\
\midrule
\texttt{get\_ci\_historical()} &
grid, date &
Hourly CI &
Get historical carbon intensity (CI) for the specified grid and date. \\
\addlinespace

\texttt{get\_ci\_forecasts()} &
grid, date, horizon, PI &
Hourly CI forecasts &
Get CI forecasts for the given horizon hours (max 96) on the specified date. Returns prediction intervals if \texttt{PI} is True. \\
\addlinespace

\texttt{get\_forecast\_accuracy()} &
grid, date, horizon &
MAPE (\%) &
Get CI forecasting accuracy for the given horizon hours (max 96) on the specified date. Returns error if ground truth is not available. \\
\addlinespace

\texttt{get\_missing\_ci\_data()} &
CI series, mask &
Imputed CI &
Get imputed CI data from the specified series. 
A mask value of 0 indicates missing data. \\
\addlinespace

\texttt{get\_supported\_grids()} &
-- &
Grid list &
List of currently supported grids. \\
\addlinespace

\texttt{set\_model()} &
model\_name, mode &
Success/ Failure &
Specify the TSFM model and mode (ZS, FT). \\
\addlinespace

\bottomrule
\end{tabularx}
\caption{\sysname API Functions to access worldwide carbon intensity data and forecasts. (ZS: Zero-Shot; FT: Fine-Tuned)}
\label{tab:carbonx-api}
\end{table*}

\section{\sysname API Suite}\label{appendix:api-suite}

Table~\ref{tab:carbonx-api} lists the APIs currently supported by \sysname that other applications can use to obtain carbon intensity data and forecasts. \sysname has a concise suite of APIs for easy integration. In the current version, downstream applications can get carbon intensity data (\texttt{get\_ci\_historical()}) and forecasts (\texttt{get\_ci\_forecasts()}) for 214 grids between the years 2021 and 2024. We are updating the implementation to provide support for fetching data and forecasts in real time. Users can also query the accuracy of forecasts from any previous date (\texttt{get\_forecast\_accuracy()}), and choose the model with the best accuracy or task requirements (\texttt{set\_model()}). Finally, users can also use \sysname to estimate any missing carbon intensity data, or get data at a finer temporal resolution than the hourly granularity if required (\texttt{get\_missing\_ci\_data()}).

\begin{table*}[t]
\centering
\label{tab_moment_combined_ciso_pjm}
\setlength{\extrarowheight}{-2pt}
\resizebox{\textwidth}{!}{
\begin{tabular}{r|ccccccccccccccccccccccccccccccc}
    \toprule
    & D1 & D2 & D3 & D4 & D5 & D6 & D7 & D8 & D9 & D10 & D11 & D12 & D13 & D14 & D15 & D16 & D17 & D18 & D19 & D20 & D21 \\
    \midrule
    \multicolumn{21}{c}{\textbf{California~(CISO) Grid}} \\ 

1*24  & 10.33 & 12.03 & 13.14 & 14.00 & 14.60 & 14.98 & 15.26 & 15.34 & 15.49 & 15.45 & 15.42 & 15.38 & 15.33 & 15.24 & 15.33 & 15.31 & 15.42 & 15.53 & 15.65 & 15.74 & 15.78 \\

2*24  &  9.52 & 11.21 & 12.39 & 13.30 & 13.82 & 14.13 & 14.33 & 14.44 & 14.46 & 14.51 & 14.37 & 14.31 & 14.24 & 14.17 & 14.23 & 14.31 & 14.43 & 14.45 & 14.62 & 14.68 & 14.82 \\

4*24  &  8.98 & 10.94 & 12.05 & 12.81 & 13.19 & 13.40 & 13.51 & 13.56 & 13.55 & 13.40 & 13.37 & 13.28 & 13.16 & 13.21 & 13.31 & 13.39 & 13.45 & 13.57 & 13.67 & 13.80 & 13.85 \\

7*24  &  8.78 & 10.44 & 11.36 & 11.93 & 12.28 & 12.34 & 12.52 & 12.58 & 12.45 & 12.38 & 12.37 & 12.38 & 12.42 & 12.42 & 12.52 & 12.61 & 12.69 & 12.80 & 12.91 & 13.02 & 13.00 \\

15*24 &  8.03 &  9.49 & 10.37 & 10.96 & 11.34 & 11.48 & 11.63 & 11.82 & 11.79 & 11.83 & 11.87 & 11.93 & 11.98 & 11.97 & 12.00 & 12.01 & 12.10 & 12.24 & 12.20 & 12.29 & 12.38 \\

30*24 &  8.01 &  9.49 & 10.27 & 10.79 & 11.16 & 11.44 & 11.62 & 11.82 & 11.88 & 11.95 & 11.89 & 12.03 & 12.06 & 12.09 & 12.24 & 12.25 & 12.35 & 12.46 & 12.55 & 12.58 & 12.70 \\









    \midrule
    \multicolumn{21}{c}{\textbf{Texas~(ERCOT) Grid}} \\ 

1*24  & 13.39 & 15.47 & 16.62 & 17.49 & 18.20 & 18.79 & 19.21 & 19.34 & 19.35 & 19.24 & 19.13 & 19.14 & 19.16 & 19.12 & 19.17 & 19.21 & 19.26 & 19.25 & 19.29 & 19.31 & 19.39 \\

2*24  & 13.55 & 15.24 & 16.41 & 17.30 & 17.94 & 18.44 & 18.61 & 18.78 & 18.73 & 18.63 & 18.51 & 18.55 & 18.48 & 18.47 & 18.54 & 18.53 & 18.59 & 18.65 & 18.70 & 18.75 & 18.80 \\

4*24  & 12.75 & 14.86 & 16.07 & 16.79 & 17.43 & 17.72 & 17.81 & 17.83 & 17.79 & 17.71 & 17.62 & 17.57 & 17.61 & 17.53 & 17.56 & 17.62 & 17.60 & 17.64 & 17.67 & 17.73 & 17.78 \\

7*24  & 12.48 & 14.58 & 15.48 & 16.17 & 16.33 & 16.71 & 16.82 & 16.80 & 16.79 & 16.66 & 16.54 & 16.61 & 16.61 & 16.65 & 16.65 & 16.70 & 16.67 & 16.71 & 16.79 & 16.69 & 16.74 \\

15*24 & 11.63 & 13.49 & 14.60 & 15.00 & 15.33 & 15.68 & 15.75 & 15.74 & 15.70 & 15.74 & 15.73 & 15.74 & 15.77 & 15.66 & 15.69 & 15.76 & 15.71 & 15.74 & 15.71 & 15.81 & 15.84 \\

30*24 & 11.55 & 13.32 & 14.23 & 14.64 & 15.11 & 15.22 & 15.39 & 15.45 & 15.49 & 15.34 & 15.38 & 15.48 & 15.42 & 15.54 & 15.49 & 15.57 & 15.67 & 15.63 & 15.72 & 15.65 & 15.70 \\

    \midrule
    \multicolumn{21}{c}{\textbf{Germany (DE) Grid}} \\ 

1*24  & 18.00 & 22.14 & 24.27 & 25.89 & 26.77 & 27.27 & 27.14 & 27.34 & 27.43 & 27.51 & 27.29 & 27.34 & 27.55 & 27.84 & 28.03 & 28.25 & 28.42 & 28.56 & 28.61 & 28.83 & 28.96 \\

2*24  & 16.79 & 21.50 & 23.66 & 25.10 & 25.72 & 25.97 & 25.92 & 26.03 & 26.13 & 25.98 & 26.04 & 26.15 & 26.42 & 26.64 & 26.83 & 26.97 & 27.16 & 27.23 & 27.41 & 27.58 & 27.78 \\

4*24  & 16.02 & 20.69 & 22.30 & 23.47 & 23.78 & 23.88 & 24.05 & 24.02 & 23.89 & 23.99 & 24.17 & 24.35 & 24.60 & 24.88 & 25.08 & 25.18 & 25.23 & 25.50 & 25.67 & 25.80 & 26.08 \\

7*24  & 15.04 & 19.13 & 20.75 & 21.49 & 22.12 & 22.25 & 22.41 & 22.57 & 22.56 & 22.97 & 23.12 & 23.40 & 23.60 & 23.72 & 24.00 & 24.05 & 24.17 & 24.50 & 24.68 & 24.80 & 25.02 \\

15*24 & 14.83 & 18.26 & 20.43 & 21.15 & 21.67 & 21.94 & 22.19 & 22.33 & 22.53 & 22.74 & 23.21 & 23.29 & 23.35 & 23.76 & 23.84 & 24.06 & 24.24 & 24.34 & 24.50 & 24.64 & 24.79 \\

30*24 & 14.17 & 18.20 & 20.07 & 21.06 & 21.82 & 21.84 & 22.24 & 22.47 & 22.69 & 22.89 & 22.93 & 23.42 & 23.50 & 23.83 & 24.02 & 24.25 & 24.36 & 24.41 & 24.70 & 24.79 & 25.07 \\

    \midrule
    \multicolumn{21}{c}{\textbf{Sweden (SE) Grid}} \\ 

1*24  & 5.03 & 6.30 & 6.94 & 7.26 & 7.55 & 7.72 & 7.86 & 7.97 & 8.13 & 8.22 & 8.29 & 8.35 & 8.43 & 8.49 & 8.59 & 8.66 & 8.76 & 8.86 & 8.92 & 8.99 & 9.00 \\

2*24  & 4.89 & 6.10 & 6.65 & 7.01 & 7.23 & 7.41 & 7.53 & 7.69 & 7.80 & 7.85 & 7.96 & 8.02 & 8.11 & 8.16 & 8.24 & 8.31 & 8.45 & 8.55 & 8.62 & 8.65 & 8.63 \\

4*24  & 4.55 & 5.70 & 6.19 & 6.47 & 6.66 & 6.83 & 7.00 & 7.16 & 7.24 & 7.42 & 7.50 & 7.56 & 7.61 & 7.72 & 7.84 & 7.95 & 8.06 & 8.14 & 8.18 & 8.22 & 8.28 \\

7*24  & 4.41 & 5.45 & 6.09 & 6.31 & 6.56 & 6.69 & 6.81 & 6.97 & 7.06 & 7.23 & 7.36 & 7.41 & 7.46 & 7.59 & 7.75 & 7.83 & 7.91 & 7.98 & 8.00 & 8.05 & 8.04 \\

15*24 & 4.26 & 5.39 & 5.97 & 6.28 & 6.56 & 6.70 & 6.89 & 7.02 & 7.12 & 7.22 & 7.32 & 7.43 & 7.52 & 7.56 & 7.67 & 7.74 & 7.84 & 7.86 & 7.91 & 7.92 & 7.91 \\

30*24 & 4.17 & 5.24 & 5.85 & 6.22 & 6.52 & 6.74 & 6.93 & 7.05 & 7.22 & 7.30 & 7.36 & 7.46 & 7.60 & 7.65 & 7.71 & 7.82 & 7.88 & 7.96 & 7.95 & 8.03 & 7.98 \\

    \bottomrule
\end{tabular}
}
\caption{\emph{Mean MAPE of \sysname (Sundial) for varying input lengths~(rows) and prediction horizons~(D1 to D21, columns) across four representative grids.}}
\label{tab:long-horizon-forecast-degradtion_appendix}
\end{table*}

\section{Zero-Shot Variant Analysis}
\label{appendix:zero-shot-variants}

We further examine \sysname zero-shot performance using alternative Sundial and Time-MoE variants with different parameter sizes.  
This analysis highlights how model scale influences forecasting accuracy.  
Table~\ref{tab:zero-shot-variants} reports 4-day MAPE for the 13 benchmark grids along with overall mean and 90th-percentile results.

Table~\ref{tab:sota-vs-tier1-zero-shot} in the main text evaluates the Sundial-Base-128M and Time-MoE-50M models.  
In contrast, Table~\ref{tab:zero-shot-variants} presents results for the Sundial-Timer-84M and Time-MoE-200M variants.

For Sundial, the 84M variant records a mean MAPE of 9.97\% and a 90th-percentile of 21.62\%.  
This is slightly higher than the 128M base model in Table~\ref{tab:sota-vs-tier1-zero-shot}, showing that additional parameters improve the modeling of fine-grained dynamics.  
Performance remains strong on stable grids such as FPL with 3.33\% MAPE, but the gap widens on more variable grids such as ERCOT with 16.19\%, where larger models capture variability more effectively.  
The Timer configuration, while efficient and competitive, highlights the trade-off between model size and forecasting accuracy.

For the Time-MoE model, the larger 200M variant achieves a mean MAPE of 9.54\% and a 90th-percentile MAPE of 20.02\%, showing slight improvement on average over the smaller 50M version in Table~\ref{tab:sota-vs-tier1-zero-shot}. The additional parameters allow the model to capture more complex temporal patterns in carbon intensity, producing more accurate zero-shot forecasts in some grids such as California (CISO) and Washington (BPAT). However, in some other grids, such as Spain (ES), accuracy decreases slightly, underscoring the need for more detailed analysis and ablation studies.

Overall, we find that scaling model size enhances zero-shot accuracy on average, with gains more pronounced for the Sundial model. 
These findings suggest that while larger configurations increase the model footprint, they can typically reduce forecasting errors. Thus, practitioners should decide on the right model variant based on their requirements and resource availability.

\section{Extended Forecasting Details}\label{appendix:long_forecasting}

Table~\ref{tab:long-horizon-forecast-degradtion_appendix} provides a detailed view of Sundial’s mean MAPE as a function of both the input (lookback) window and the forecast horizon across four representative grids: California~(CISO), Texas~(ERCOT), Germany~(DE), and Sweden~(SE).

Longer lookback windows provide better accuracy at shorter horizons. In every grid, a 30-day input consistently yields the lowest error for day-ahead (D1) forecasts. For example, CISO records 8.01\% MAPE at D1 with a 30-day history compared with 10.33\% when the input is just the past day. Similar patterns appear in ERCOT at 11.55\% versus 13.39\%, DE at 14.17\% versus 18.00\%, and SE at 4.17\% versus 5.03\%,
showing that long histories capture richer context for short-term prediction.

As the horizon extends, accuracy decreases regardless of the lookback window length. Similar to the case of shorter horizons, longer lookbacks achieve lower errors. However, interestingly, for most grids, a 15-day lookback window starts achieving better accuracy than a 30-day window from day 7 onwards, as evident in CISO, DE, and SE. For example, while day four forecasting error in CISO is 10.96\% with a 15-day lookback compared to 10.79\% with a 30-day lookback, day 8 errors are the same, and at day 21, a 15-day lookback has a 12.38\% error compared to 12.70\% in the other case. These results confirm that, while additional historical context helps capture slow seasonal patterns needed for multi-week forecasts, excessively longer windows may amplify the noise in the data.

\begin{table*}[t]
\centering
\begin{threeparttable}
\resizebox{0.85\columnwidth}{!}{
\begin{tabular}{l cc| cc}
\toprule
& \multicolumn{4}{c}{\textbf{\sysname Zero-Shot Mode}} \\
\cmidrule(lr){2-5}
& \multicolumn{2}{c}{\text{Sundial~\cite{liu2025sundial}}~(84M)} &
  \multicolumn{2}{c}{\text{Time-MoE~\cite{shi2024time}}~(200M)} \\
\cmidrule(lr){2-3}
\cmidrule(lr){4-5}
& Mean & 90th & Mean & 90th \\
\midrule
CISO     & 11.35 & 24.14 & 11.12 & 23.35 \\
         
PJM      &  4.87 & 10.41 &  4.69 &  10.42 \\

ERCOT    & 16.19 & 35.08 &  15.53 & 35.82 \\
         
ISO-NE   & 6.96 & 14.81 &  6.75 & 14.24 \\
         
BPAT     & 9.88 & 20.11 & 9.64 & 19.75 \\
         
FPL      &  3.33 & 7.72 &  2.77 &  6.26 \\
         
NYISO    & 9.02 & 19.55 &  7.64 & 15.30 \\
         
SE       & 6.29 & 14.67 &  6.52 & 14.17 \\
         
DE       & 20.71 & 47.99 & 20.44 & 41.23 \\
         
PL       &  6.90 & 14.87 &  6.32 & 13.36 \\
         
ES       & 21.26 & 44.53 & 20.13 & 40.03 \\
         
NL       & 9.19 & 18.72 &  8.93 & 17.97 \\
         
AUS-QLD  & 3.68 & 8.44 &  3.53 &  8.35 \\
         
\midrule
\textbf{Average}
         & 9.97 & 21.62 &  9.54 & 20.02 \\
\bottomrule
\end{tabular}
}
\caption{\emph{4-day zero-shot MAPE for \sysname using Sundial and Time-MoE with different variants.}}
\label{tab:zero-shot-variants}
\end{threeparttable}
\end{table*}

\end{document}